\pdfoutput=1

\documentclass[11pt]{article}

\usepackage[]{ACL2023}

\usepackage{times}
\usepackage{latexsym}

\usepackage[T1]{fontenc}

\usepackage[utf8]{inputenc}

\usepackage{microtype}

\usepackage{inconsolata}

\usepackage{graphicx}
\usepackage{colortbl}
\usepackage{amsfonts}
\usepackage{multirow}
\usepackage{color}
\usepackage{subfigure}
\usepackage{xcolor}
\usepackage{booktabs}       
\usepackage{amsmath}
\usepackage{times}
\usepackage{textcomp}
\usepackage{enumitem}
\definecolor{yel}{RGB}{250,247,228}
\definecolor{gray}{RGB}{195,195,195}

\title{NoisywikiHow: A Benchmark for Learning with Real-world Noisy Labels in Natural Language Processing}

\author{Tingting Wu$^1$, 
Xiao Ding$^1$\thanks{~~Corresponding author.}, 
Minji Tang$^1$, 
Hao Zhang$^2$,
Bing Qin$^1$, 
Ting Liu$^1$ \\
$^1$\normalsize{Research Center for Social Computing and Information Retrieval}\\[-.05cm]
\normalsize{Harbin Institute of Technology, China}\\[-.05cm]
{\small\tt\{ttwu, xding, mjtang,qinb tliu\}@ir.hit.edu.cn}\\[-.05cm]
$^2$\normalsize{Faculty of Computing, Harbin Institute of Technology, China}\\[-.05cm]
{\small\tt zhh1000@hit.edu.cn}}

\begin{document}
\maketitle
\begin{abstract}
Large-scale datasets in the real world inevitably involve label noise. Deep models can gradually overfit noisy labels and thus degrade model generalization. 
To mitigate the effects of label noise, learning with noisy labels (LNL) methods are designed to achieve better generalization performance. 
Due to the lack of suitable datasets, previous studies have frequently employed synthetic label noise to mimic real-world label noise. 
However, synthetic noise is not instance-dependent, making this approximation not always effective in practice. Recent research has proposed benchmarks for learning with real-world noisy labels. 
However, the noise sources within may be single or fuzzy, making benchmarks different from data with heterogeneous label noises in the real world. 
To tackle these issues, we contribute \emph{NoisywikiHow}, the largest NLP benchmark built with minimal supervision. 
Specifically, inspired by human cognition, we explicitly construct multiple sources of label noise to imitate human errors throughout the annotation, replicating real-world  noise, whose corruption is affected by both ground-truth labels and instances. 
Moreover, we provide a variety of noise levels to support controlled experiments on noisy data, enabling us to evaluate LNL methods systematically and comprehensively. 
After that, we conduct extensive multi-dimensional experiments on a broad range of LNL methods, obtaining new and intriguing findings.\footnote{The dataset is publicly available at  \url{https://github.com/tangminji/NoisywikiHow}.}
\end{abstract}

\section{Introduction}
Large-scale labeled data has become indispensable in the notable success of deep neural networks (DNNs) in various domains and tasks~\citep{russakovsky2015imagenet,wang2018glue}. 
Due to imperfect sources like crowd-sourcing and web crawling~\citep{xiao2015learning,zhang2017position,lee2018cleannet}, datasets frequently include \emph{real-world label noise}~\citep{chen2021beyond}, which may induce model overfitting to noisy labels and hurt the generalization of deep models~\citep{ChiyuanZhang2017UnderstandingDL,wu2022stgn,wu2022discrimloss}. 
To alleviate this issue, learning with noisy labels (LNL) methods for robustly training deep models have been studied extensively.
\begin{table}[]
\small\setlength\tabcolsep{1.5pt}
\centering
\begin{tabular}{llc}
\toprule
\multicolumn{2}{c}{\textbf{Input}}                                                                        & \textbf{Output}                \\ \midrule
(a) & Take prescription weight loss medications.                                                 & \multirow{4}{*}{\begin{tabular}[c]{@{}c@{}}Losing\\ Weight\end{tabular}} \\ \cmidrule{0-1}
(b) & Check calories on food packaging.                                                          &                                \\ \cmidrule{0-1}
(c) & Include cultural and ethnic foods in your plan. &                                \\ \cmidrule{0-1}
(d) & Talk about food differently.                                                               &                                \\ \bottomrule
\end{tabular}
\caption{Instances (a)–(d) depict examples of our task. \textbf{Input}: a procedural event. \textbf{Output}: a plausible intention toward that event.}
\label{example_intro}
\end{table}

Due to the lack of appropriate benchmarks, previous research often studied synthetic label noise to simulate real-world label noise~\citep{zhang2018mixup,lukasik2020does}. As a general and realistic noise, real-world noise may have several noise sources (i.e., be \emph{heterogeneous})~\citep{northcutt2021pervasive} and be \emph{instance-dependent} (i.e., $P(\tilde{y}|y,x)$, where the probability of an instance being assigned to the incorrect label $\tilde{y}$ depends on the original ground-truth label $y$ and data $x$)~\citep{han2020survey,song2022learning}. However, synthetic noise is generated from an artificial distribution and is thus \emph{instance-independent} (i.e., $P(\tilde{y}|y)$), which may not always work well in practice.


Recently, various benchmarks for learning with real-world noisy labels have been proposed across fields like computer vision (CV)~\citep{li2017webvision}, audio signal processing (ASP)~\citep{gemmeke2017audio}, and natural language processing (NLP)~\citep{hedderich2021analysing}. 
To fully evaluate robust learning methods with real-world label noise, benchmarks should be as close to real-world scenarios as possible. 
Meanwhile, controlled experiments are encouraged to verify whether LNL methods can remain effective over a wide range of noise levels~\citep{jiang2020beyond}. 
Nevertheless, the noise levels in most datasets are fixed and unknown, resulting in uncontrolled label noise~\cite{fonseca2019learning,song2019selfie}. 
Moreover, the noise therein often comes from the same or ambiguous sources~\citep{li2017webvision,jiang2020beyond}, which conflicts with the heterogeneous characteristics of real-world noise. 
These problems prevent a better understanding of LNL methods.

To bridge this gap, we present NoisywikiHow, a new NLP benchmark for evaluating LNL methods focusing on the intention identification task.  
Intention identification promotes numerous downstream natural language understanding tasks, from commonsense reasoning~\citep{sap2019atomic} to dialogue systems~\citep{pepe2022steps}. 
Additionally, the complexity of the task (total of 158 categories) facilitates a deeper investigation of the efficacy of LNL approaches. 
The task form is shown in Table~\ref{example_intro}.
 
To make the benchmark more representative of real-world scenarios, we propose a practical assumption: \emph{Real-world label noise in a dataset is mainly induced by human errors, regardless of whether the dataset's construction is automated or crowd-sourced.}
Existing psychological and cognitive evidence further supports our hypothesis. It shows that different annotators have different preferences and biases~\citep{beigman2009learning,burghardt2018quantifying}, 
which means human labeling errors typically result from multiple noise sources. 
Furthermore, humans may make random labeling errors due to random attention slips. 
But they are more likely to produce label noise when labeling hard cases~\citep{klebanov2008analyzing} (i.e., noise is instance-dependent), such as instance (c) in Table~\ref{example_intro}. 


Motivated by this human cognition, 
we first collect data from the wikiHow website,\footnote{\url{https://www.wikihow.com}} which contains a collection of professionally edited how-to guideline articles, providing a vast quantity of clean scripts and corresponding categories for free to help achieve controlled experiments and ensure benchmark quality. 
After that, we explicitly inject a variety of noise sources into clean data to replicate human annotation errors, thus introducing real-world label noise into the benchmark.
Notably, training samples in our benchmark exhibit a long-tailed class distribution,  which is in line with the facts, i.e., data in real-world applications is heavily imbalanced~\citep{van2018inaturalist,liu2019large}. 
Besides, we achieve minimal human supervision by using a series of automated labeling procedures, saving lots of time and human effort.

To evaluate NoisywikiHow, we carry out extensive experimentation across various model architectures and noise sources, execute plentiful LNL methods on our benchmark, compare the more realistic real-world noise with the extensively studied synthetic noise, and investigate a case study and long-tailed distribution characteristics. 

\section{Related Work}
\subsection{Datasets with real-world noisy labels}

\begin{table*}[]
\small
\centering
\begin{tabular}{lccccc}
\toprule
\textbf{Dataset}           & \textbf{Classes} & \textbf{Distribution} & \textbf{Controlled} & \textbf{\begin{tabular}[c]{@{}c@{}}Human\\ Annotation\end{tabular}} & \textbf{Size}  \\ \midrule
\rowcolor{yel} \multicolumn{6}{l}{CV}                                                                                \\
Food-101N~\citep{lee2018cleannet}         & 101  & Balanced    & No    & No                                                    & 367K  \\
Animal-10N~\citep{song2019selfie}        & 10   & Balanced    & No    & Yes                                                    & 55K   \\
Red MiniImageNet~\citep{jiang2020beyond}  & 100  & Balanced    & Yes    & Yes                                                    & 55K   \\
Red Stanford Cars~\citep{jiang2020beyond} & 196  & Balanced    & Yes    & Yes                                                    & 16.1K \\
Clothing1M~\citep{xiao2015learning}        & 14   & Imbalanced    & No    & No                                                    & 1M    \\
WebVision~\citep{li2017webvision}         & 1K   & Long-tailed   & No    & No                                                    & 2.4M  \\ \midrule
\rowcolor{yel} \multicolumn{6}{l}{ASP}                                                                             \\
AudioSet~\citep{gemmeke2017audio}          & 527  & Long-tailed   & No    & Yes                                                    & 2M    \\
FSDnoisy18K~\citep{fonseca2019learning}       & 20   & Imbalanced    & No    & No                                                    & 18.5K \\
FSDKaggle2019~\citep{fonseca2019audio}     & 80   & Balanced    & No    & No                                                    & 29.2K \\ \midrule
\rowcolor{yel} \multicolumn{6}{l}{NLP}                                                                               \\
NoisyNER~\citep{hedderich2021analysing}          & 4    & Long-tailed   & Yes    & No                                                    & 14.8K \\ \midrule
\textbf{NoisywikiHow}      & \textbf{158}  & \textbf{Long-tailed}   & \textbf{Yes}    & \textbf{No}                                                    & \textbf{89K}   \\ \bottomrule
\end{tabular}
\caption{\label{comp_datasets}
Comparison between our benchmark and other datasets. 
}
\end{table*}

In early studies of the LNL problem, due to a lack of appropriate benchmarks, synthetic noise was often used to reflect noise in the real world and assess the effectiveness of methods~\citep{BoHan2018CoteachingRT,zhang2018mixup}. However, unlike real-world noise, synthetic noise follows an idealized artificial distribution, which leads to inaccurate approximations and inadequate evaluations.

Recent studies have proposed numerous datasets with real-world noisy labels. Table~\ref{comp_datasets} depicts a  comparison of existing real-world noisy datasets for evaluating LNL methods in CV, ASP, and NLP. As shown in Table~\ref{comp_datasets}, most datasets fail to perform controlled experiments on real-world label noise and cannot be used to study DNNs across different noise levels~\citep{fonseca2019learning,fonseca2019audio}.

A few benchmarks with controlled label noise, such as NoisyNER~\cite{hedderich2021analysing} and Red MiniImageNet~\cite{jiang2020beyond}, were produced. 
However, the noise source in their datasets may be vague.  
Furthermore, NoisyNER focuses on the named entity recognition task in NLP. Though seven noisy label sets are provided, it is challenging to determine the precise noise level of each label set because a sentence-level instance has numerous word-level labels. 
Besides, Red MiniImageNet relies heavily on careful human annotation and follows a balanced class distribution, which diverges from real-world application scenarios. In this paper, we publish NoisywikiHow to solve the above limitations. As shown in Table~\ref{comp_datasets}, to the best of our knowledge, NoisywikiHow is the largest NLP benchmark for assessing LNL methods.



\subsection{Intention identification}

Intention identification is critical to many applications~\cite{huang2016liberal,sap2019atomic}. Therefore, ensuring task reliability is essential. 
Some previous work formulates intention identification as an \emph{event process typing} task. 
Given a sequence of events, the model is designed to understand the overall goal of the event process in terms of an action and an object~\citep{chen2020you,pepe2022steps}.
In other studies, intention identification is modeled as a \emph{sentence classification} task~\citep{zhang2020intent,zhang2020reasoning}. 
When given a procedural event, the system predicts its intention in a 4-choose-1 multiple-choice format. 
However, none of these studies deal with task reliability. By building NoisywikiHow, we make a preliminary exploration of  task reliability (i.e., model performance under label noise). 
Following~\citet{zhang2020reasoning}, we model intention identification as a sentence classification task. The difference is that our benchmark (including 158 labels) is analogous to the \emph{retrieval task} in a more practical and challenging way.

\section{NoisywikiHow Dataset}
\subsection{Data Collection}\label{data_collection} 
We construct NoisywikiHow by crawling how-to articles from the wikiHow website. 
Detailed crawling strategies and related statistics are in Appendix~\ref{app:crawling}. 
We define the \textbf{input} as a procedural event, i.e., the header of a paragraph in a wikiHow article (e.g., \emph{Talk about food differently} in Table~\ref{example_intro}), and the \textbf{output} as the intention of the event, namely the category of this article (e.g., \emph{Losing Weight} in Table~\ref{example_intro}).
Note that categories present a hierarchy (e.g., \emph{Health \(\gg\) Nutrition and Food Health \(\gg\) Weight Management \(\gg\) Losing Weight}), and we select the category with the finest granularity as the label.

\begin{figure}[t]
\centering
\includegraphics[width=1.0\columnwidth]{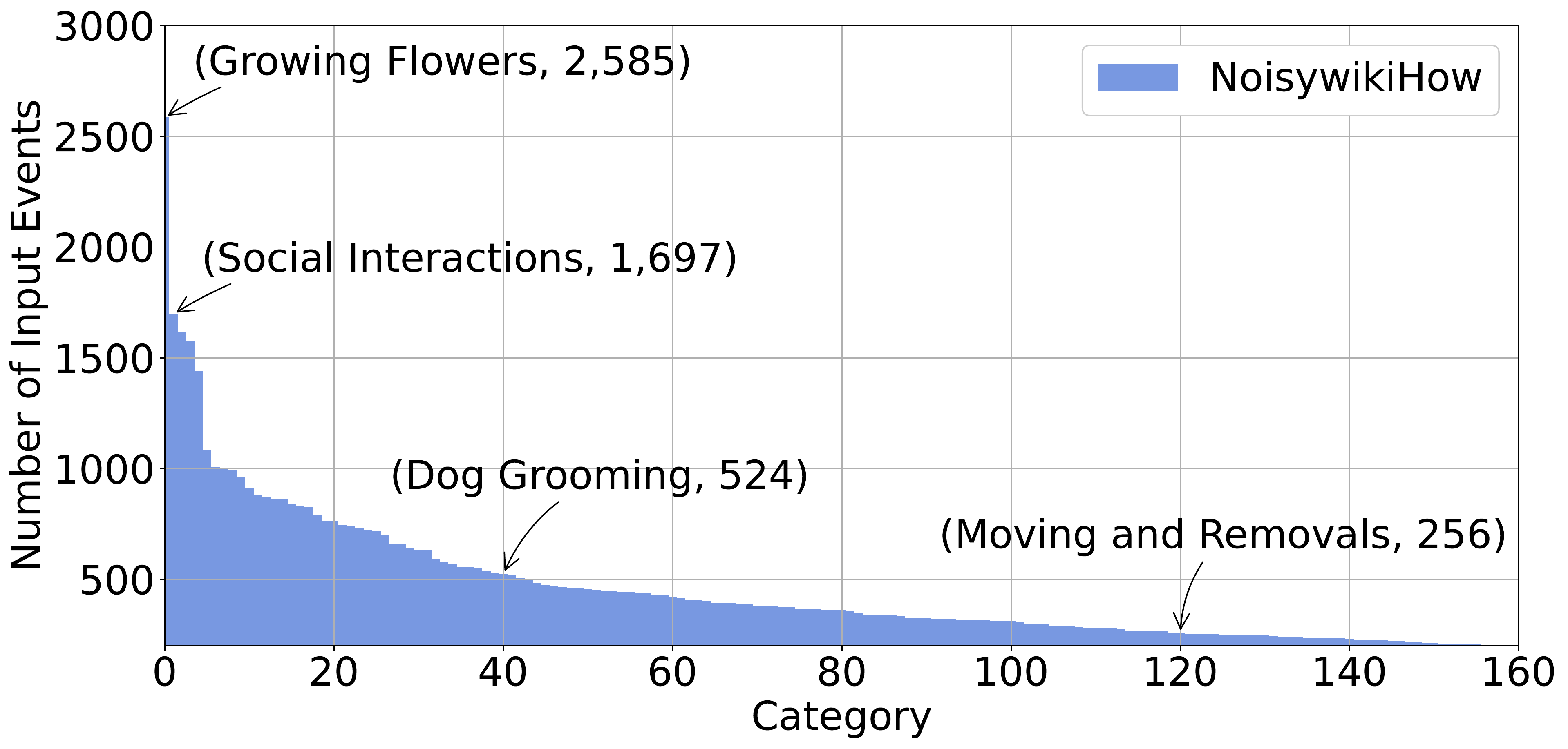}
\caption{Number of events per category of the training set of NoisywikiHow.}
\label{lt_dis}
\end{figure}
\subsection{Data Cleaning}\label{data_cleaning}
Similar to~\citet{jiang2020beyond} and~\citet{hedderich2021analysing}, we realize controlled label noise by injecting various amounts of noise into clean data. However, the data collection process introduces a lot of low-quality or irrelevant data. As a result, we develop a data cleaning procedure to remove bad data and facilitate the target task from two aspects: (1) input filtering and (2) label filtering.
Regarding input filtering, we first devise four automatic filters and execute them sequentially to remove low-quality or ambiguous data.
\begin{itemize}[itemsep=0pt, topsep=3pt, partopsep=3pt]
\item \textbf{Sample Length Filter} intends to retain instances with more informative and complete semantic information by filtering excessively short or long data. 
\item \textbf{Format Normalization} is to standardize instances (e.g., unifying the description of ``\emph{Click Defragment \textbf{Y}our \textbf{H}ard \textbf{D}rive.}'' and ``\emph{Click Defragment \textbf{y}our \textbf{h}ard \textbf{d}rive.}''), ensuring the effectiveness of subsequent strategies.
\item \textbf{Deduplication} tries to eliminate redundant or ambiguous data (e.g., a procedural event corresponds to multiple intents).
\item \textbf{TF-IDF Filter} attempts to preclude overly uninformative instances by calculating the TF-IDF for each token.
\end{itemize}


After that, we receive \textbf{\emph{high-quality data} $\mathcal{D}_h$}, which follows a long-tailed class distribution with limited data on tail classes, resembling the distribution in Fig.~\ref{lt_dis}. We create a \textbf{Sample Size Filter} to exclude the categories with too few samples ($\leq300$), ensuring an appropriate split of training, validation, and test sets.


We observe that the labels have two types, i.e., concepts defined as nominal phrases (e.g., \emph{Nutrition and Food Health}), and event mentions defined as nominal or verbal phrases that refer to events (e.g., \emph{Losing Weight})~\cite{min2020towards,yu2021event}. 
Therefore, label filtering is required to retain only events, ensuring the effectiveness of intention identification. 
Specifically, each category is annotated by three graduate students from the NLP field and is regarded as an event if more than two annotators agree. Human annotators are asked to label 736 categories and achieve a high agreement  (Fleiss-$\kappa$ = 0.84)~\citep{fleiss1971measuring}. 

After data cleaning, we obtain \textbf{\emph{clean data} $\mathcal{D}$} involving 89,143 instances in 158 classes. Due to the limited space, complete filtering strategies and more details are in Appendix~\ref{app:filtering}.


\subsection{Label Noise Injection}
To create a benchmark of real-world noisy labels, we introduce various sources of controlled label noise into the clean data. Prior to this, we assume that \emph{\textbf{human mistake is the primary cause of real-world label noise in a dataset}}. 
Psychological and cognitive findings further corroborate the rationality of the assumption. It demonstrates that: 
(1) apparent differences between annotators result from different preferences and biases~\citep{reidsma2008exploiting,beigman2009learning,burghardt2018quantifying}, suggesting that human errors are \emph{\textbf{heterogeneous}}; 
(2) label noise from humans regularly affects hard cases~\citep{klebanov2008analyzing,klebanov2009squibs}, proving that noise is \emph{\textbf{instance-dependent}}. 

\noindent\textbf{Heterogeneous noise sources.}\quad
Based on the above preliminaries, we simulate various mistakes committed by annotators to produce real-world noise containing heterogeneous noise sources. Specifically, human errors are often induced by ambiguity, insufficient annotator expertise, and random attention slips~\citep{beigman2009learning,hollenstein2016inconsistency}. Motivated by this, we develop three noise sources as follows:
\begin{itemize}[itemsep=0pt, topsep=3pt, partopsep=3pt]
\item Sub-categories (\textbf{SC}) under the same category (e.g., \emph{Starting a Business} and \emph{Running a Business}) tend to have higher semantic similarities and can be easily confused. SC depicts the noise caused by labeling ambiguous instances.
\item Intents beyond the commonsense categories (\textbf{BCC}) are hard to identify (e.g., \emph{Dog Grooming}), readily inducing noisy labels. BCC portrays a scenario annotated by a human lack of expert knowledge.
\item Considering the long-tailed distribution, even a few labeling errors on tail classes can seriously affect learning of these categories. Therefore, achieving robust training on tail classes is critical. We concentrate on intents under the tail categories (\textbf{TC}), which describe the noise generated by humans randomly shifting their attention.
\end{itemize}



Then, we design a simple mapping from noise sources to classes to facilitate the subsequent injection of noise from different sources and categories. Specifically, each class is associated with a noise source, and classes under various noise sources do not overlap. This mapping can cover all categories during noise injection and determine the potential noise source for each class. Finally, we divide 158 categories into 68, 36, and 54 to correspond to the sources SC, BCC, and TC, respectively.
More details about the mapping can be found in Appendix~\ref{app:data_analysis}.

\noindent\textbf{Injecting instance-dependent label  noise.}\quad
Since each noise source contains a set of categories, each of which may involve hard cases, instance-dependent label noise exists in each noise source. Note that real-world label noise always comes from an open rather than a finite category set~\citep{wang2018iterative}. We therefore enable label noise to derive from categories other than the current label set. However, this operation changes the number of labels and impacts the target classification task. To solve this problem, when injecting label noise into an instance $(x, y)$, we leave the label $y$ (\emph{output}) unchanged like~\citet{jiang2020beyond} but replace the procedural event $x$ (\emph{input}) with the one ($\tilde{x}$) under the other category ($\tilde{y}$), which may not be in the existing 158 classes. 
Moreover, NoisywikiHow supports five noise levels (i.e., 0\%, 10\%, 20\%, 40\%, and 60\%). Like~\citet{li2017webvision} and~\citet{saxena2019data}, we assume that, given a specified noise level $t$, $t$ is uniform across noise sources. For example, $t=10\%$ represents that each source has roughly 10\% label noise.

We further identify hard cases and inject instance-dependent noise for each noise source. 
Intuitively, when we mislabel an instance from $(x,y)$ into $(\tilde{x},y)$, if $(x,y)$ is a hard case, the semantic representations of events $x$ and $\tilde{x}$ should be very similar. As a result, for any $(x,y)$, 
we can assess its difficulty by finding an $(\tilde{x},\tilde{y})$ whose $\tilde{x}$ has the maximum semantic similarity with $x$. 
To identify $(\tilde{x},\tilde{y})$, we take the following steps: (1) \textbf{Determine $\mathcal{D}_n$: the candidate set of $(\tilde{x},\tilde{y})$}. 
To avoid introducing bad data or duplicate data after noise injection, we construct $\mathcal{D}_n$ as follows: 
\begin{itemize}[itemsep=0pt, topsep=3pt, partopsep=3pt]
\item For the sources BCC and TC,  $\mathcal{D}_n=\mathcal{D}_h-\mathcal{D}$. 
\item For the source SC, let $\mathcal{D}_s$ be the sample set of all other sub-categories except $y$ under the same category, and $\mathcal{D}_n=(\mathcal{D}_h-\mathcal{D})\cap\mathcal{D}_s$.
\end{itemize}
(2) \textbf{Locate $\tilde{x}$ in $\mathcal{D}_n$}. 
Following~\citet{zhang2020reasoning}, we map each event to a vector representation by taking the average of the BART embeddings~\citep{lewis2020bart} of the verbs. $\tilde{x}$ thus can be calculated as:
\begin{equation}
\tilde{x} = \arg\max_{v_{x'}} {cosine(v_x,v_{x'})}, (x',y')\in \mathcal{D}_n,
\end{equation} 
where $v_{(\cdot)}$ is the vector representation of an event, and  $cosine(\cdot)$ denotes the cosine similarity of two vectors.
For any $(x,y)$, its difficulty can be obtained by calculating a score $s_{x}$:
\begin{equation}\label{s_score}
s_{x}=cosine(v_x,v_{\tilde{x}}).
\end{equation}
The larger the $s_{x}$, the harder the instance $(x,y)$.

We inject noise into the training set $\mathcal{D}_{tr} \subset \mathcal{D}$. Given a specified noise level $t$ (e.g., 10\%), all instances in $\mathcal{D}_{tr}$ are arranged in decreasing order of $s_{x}$, with the top $t$ of the samples in each source considered hard cases. 
We inject instance-dependent noise by replacing $x$ for each hard case with $\tilde{x}$.
\begin{table}[]
\small\setlength\tabcolsep{4.5pt}
\begin{tabular}{cccllc}
\toprule
\multicolumn{6}{c}{\textbf{Noise Level(\%): 0, 10, 20, 40, 60}}                              \\ \midrule
\textbf{Noise Sources} & \textbf{Class} & \textbf{Train}  & \multicolumn{1}{c}{\textbf{Val}} & \multicolumn{1}{c}{\textbf{Test}} & \textbf{Total}  \\ \midrule
SC            & 68    & 39,674 & 3,400                   & 3,400                    & 46,474 \\
BCC           & 36    & 20,413 & 1,800                   & 1,800                    & 24,013 \\
TC            & 54    & 13,256 & 2,700                   & 2,700                    & 18,656 \\ \midrule
Total         & 158   & 73,343 & 7,900                   & 7,900                    & 89,143 \\ \bottomrule
\end{tabular}
\caption{\label{dataset_statistic}
Overview of NoisywikiHow of multiple noise sources and controlled label noise, where SC, BCC, and TC denote noise sources from sub-categories, categories beyond the commonsense, and tail categories.
}
\end{table}

\begin{table*}[htbp!]
\small
\centering
\begin{tabular}{cccccc}
\toprule
\multirow{3}{*}{\textbf{Method}} & \multicolumn{5}{c}{\textbf{Noise Level}}                                           \\ \cmidrule{2-6} 
& 0\%           & 10\%          & 20\%          & 40\%          & 60\%          \\ \cmidrule{2-6} 
& Top-1(Top-5)   & Top-1(Top-5)   & Top-1(Top-5)   & Top-1(Top-5)   & Top-1(Top-5)   \\ \midrule
BERT~\citep{JacobDevlin2018BERTPO}&60.29(83.53)&58.86(83.82)&57.42(82.57)&52.91(79.84)&48.20(75.37) \\
XLNET~\citep{ZhilinYang2019XLNetGA}&59.77(85.24)&60.23(85.90)&58.25(84.29)&53.74(81.73)&50.23(\textbf{79.44}) \\
RoBERTa~\citep{YinhanLiu2019RoBERTaAR}&60.59(85.10)&59.65(84.16)&57.77(83.77)&54.18(81.56)&\textbf{50.85}(78.87) \\
GPT2~\citep{radford2019language}&59.84(85.39)&58.35(84.90)&57.0(83.94)&52.71(80.81)&48.25(78.08) \\
ALBERT~\citep{lan2020albert}&55.13(80.80)&56.21(82.15)&53.68(80.52)&49.93(78.44)&44.81(74.41) \\
T5~\citep{raffel2020exploring}&58.35(83.63)&56.87(83.03)&56.19(82.20)&52.29(79.94)&47.47(77.39) \\
BART~\citep{lewis2020bart}&\textbf{61.72(86.90)}&\textbf{60.28(85.92)}&\textbf{58.94(84.67)}&\textbf{54.57(82.38)}&49.75(78.84) \\ \bottomrule
\end{tabular}
\caption{\label{dim1}
Top-1 (Top-5) classification accuracy (\%) of pre-trained language models on the NoisywikiHow test set under different levels of real-world label noise. Top-1 results are in bold.
}
\end{table*}
\section{Experiments}
We first present the general settings for experiments (Section~\ref{exp_setting}). Further, we systematically evaluate our benchmark with varied model architectures (Section~\ref{exp_model_arch}) and noise sources (Section~\ref{exp_noise_source}). 
Also, we assess a broad range of LNL methods on NoisywikiHow (Section~\ref{exp_lnl}) and compare real-world noise with synthetic noise (Section~\ref{exp_rw_vs_syn}). Finally, we 
conduct a case study
 (Section~\ref{exp_quality}). 
In addition, we discuss the long-tailed distribution characteristics of NoisywikiHow in Appendix~\ref{app:exp_lt}.


\subsection{Experimental settings}\label{exp_setting}
On our benchmark, all methods are trained on the noisy training sets\footnote{Synthetic noise and various noise sources under real-world noise correspond to diverse noisy training sets.} 
and evaluated on the same clean validation set to verify whether these approaches can resist label noise during training and achieve good generalization on the noise-free data. 
Before adding label noise, we randomly split out 15,800 instances from clean data and then equally divide them into two sets: a validation set and a test set. 
The remaining 73,343 instances serve as the training set, which follows a typical long-tailed class distribution and is analogous to heavily imbalanced data in real-world applications, as shown in Fig.~\ref{lt_dis}. The statistics of NoisywikiHow are shown in Table~\ref{dataset_statistic}. 
We cast intention identification as a classification problem. We exploit the cross-entropy loss for training models and use Top-1 accuracy and Top-5 accuracy as the evaluation metrics. 
\subsection{Comparison of Model Architectures}\label{exp_model_arch}
\textbf{Baselines}:
We first evaluate the performance of different model architectures under varying levels of real-world label noise. 
Regarding the model architectures, we use seven 
state-of-the-art (SOTA) pre-trained language models, including BERT~\citep{JacobDevlin2018BERTPO}, XLNet~\citep{ZhilinYang2019XLNetGA}, RoBERTa~\citep{YinhanLiu2019RoBERTaAR}, GPT2~\citep{radford2019language}, ALBERT~\citep{lan2020albert}, T5~\citep{raffel2020exploring}, and BART~\citep{lewis2020bart}. 
We finetune each model for 10 epochs with batch size 64, learning rate 3e-5. 
These hyperparameters remain unchanged in subsequent experiments unless indicated otherwise.
In this paper, we conduct all experiments utilizing the base-sized version of the pre-trained language models.
Besides, due to long output sequences in partial categories, we adopt beam search~\citep{sutskever2014sequence} in T5, with a beam width of 5 and a length penalty of $\alpha$ = 1.0.

\textbf{Results}:
As shown in Table~\ref{dim1}, the Top-1 accuracies of SOTA pre-trained language models on our benchmark are generally not high, and an increase in noise levels can lead to considerable performance degradation for a given model, demonstrating the challenge of the NoisywikiHow dataset. 

In Table~\ref{dim1}, different architectures are representative of diverse capacities. 
For example, RoBERTa and XLNet consistently outperform ALBERT under different noise levels. 
In addition, we observe that BART achieves the best performance among these SOTA models under a majority of noise levels, regardless of Top-1 or Top-5 classification accuracy. 
This is mainly because a better \emph{denoising} objective (i.e., \emph{masked language modeling}) is used during pre-training of BART. 
In pre-training, BART gains better denoising ability by corrupting text with an arbitrary noise function (thus making the noise more flexible) and learning to reconstruct the original text. 
In the following, we use the BART model as the \emph{\textbf{base model}}.


\subsection{Effects of Distinct Noise Sources}\label{exp_noise_source}

\begin{table}[]
\small\setlength\tabcolsep{15pt}
\centering
\begin{tabular}{cll}
\toprule
\textbf{Noise Sources} & \multicolumn{1}{c}{\textbf{Top-1}} & \textbf{Top-5}  \\ \midrule
SC+BCC+TC  & 60.28                    & 85.92 \\
SC            & 60.14                    & 85.49 \\
BCC           & 59.65                    & 85.39 \\
TC            & 57.99                    & 84.37 \\
 \bottomrule
\end{tabular}
\caption{Test accuracy (\%) of the base model under distinct noise sources with 10\% label noise, where SC+BCC+TC denotes the default NoisywikiHow with a mixture of noise sources.
}
\label{dif_noise_source}
\end{table}

We further explore the characteristics of different noise sources. To this end, we pick the same model (i.e., the base model) and separately validate the performances on individual noise sources under the same noise level. For convenience, we denote noise-free data by \emph{correct samples} and data with label noise by \emph{incorrect samples}. 

\textbf{Results}:
Table~\ref{dif_noise_source} shows the results of the base model under four different noise sources with 10\% label noise. 
As shown in Table~\ref{dif_noise_source}, there exists an evident gap between the results under noise source TC and those in other conditions. 
The label noise from noise source TC is more difficult to mitigate than others at the same noise level, mainly due to the limited data on tail categories. 
When all noisy labels are derived from TC, fewer correct samples are left, leading to inadequate model training and degradation of model performance. 
It indicates that resisting label noise from different sources may have varying difficulty levels, although the noises in these sources are all real-world label noise. Additional details are provided in Appendix ~\ref{app:exp_noise_source}.

\begin{figure*}[t]
	\centering
	\subfigure[Real-world label noise.]{
		\includegraphics[width=0.9\columnwidth]{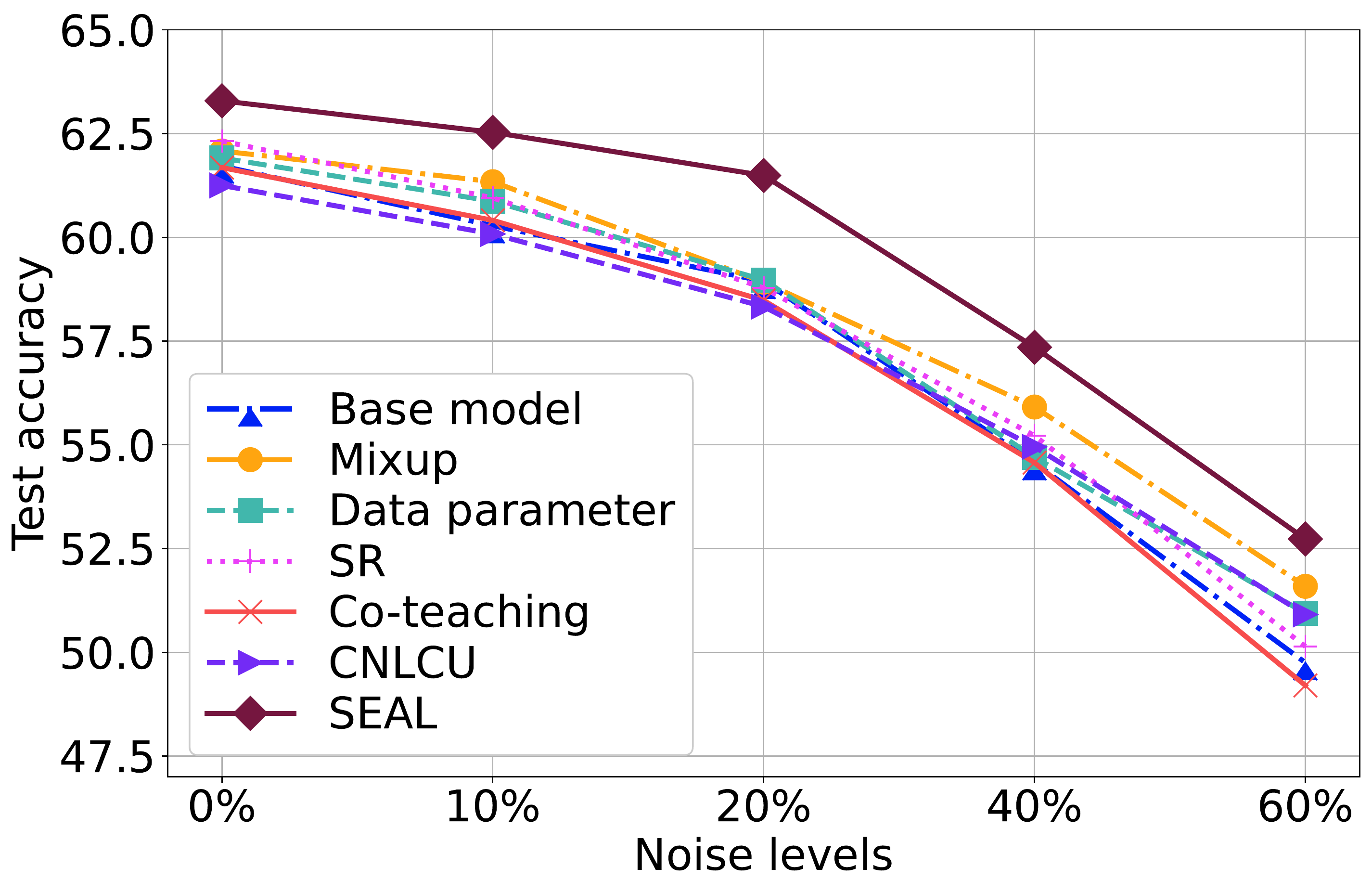}
		\label{lnl_on_idn} 
	}
	\subfigure[Synthetic label noise]{
		\includegraphics[width=0.9\columnwidth]{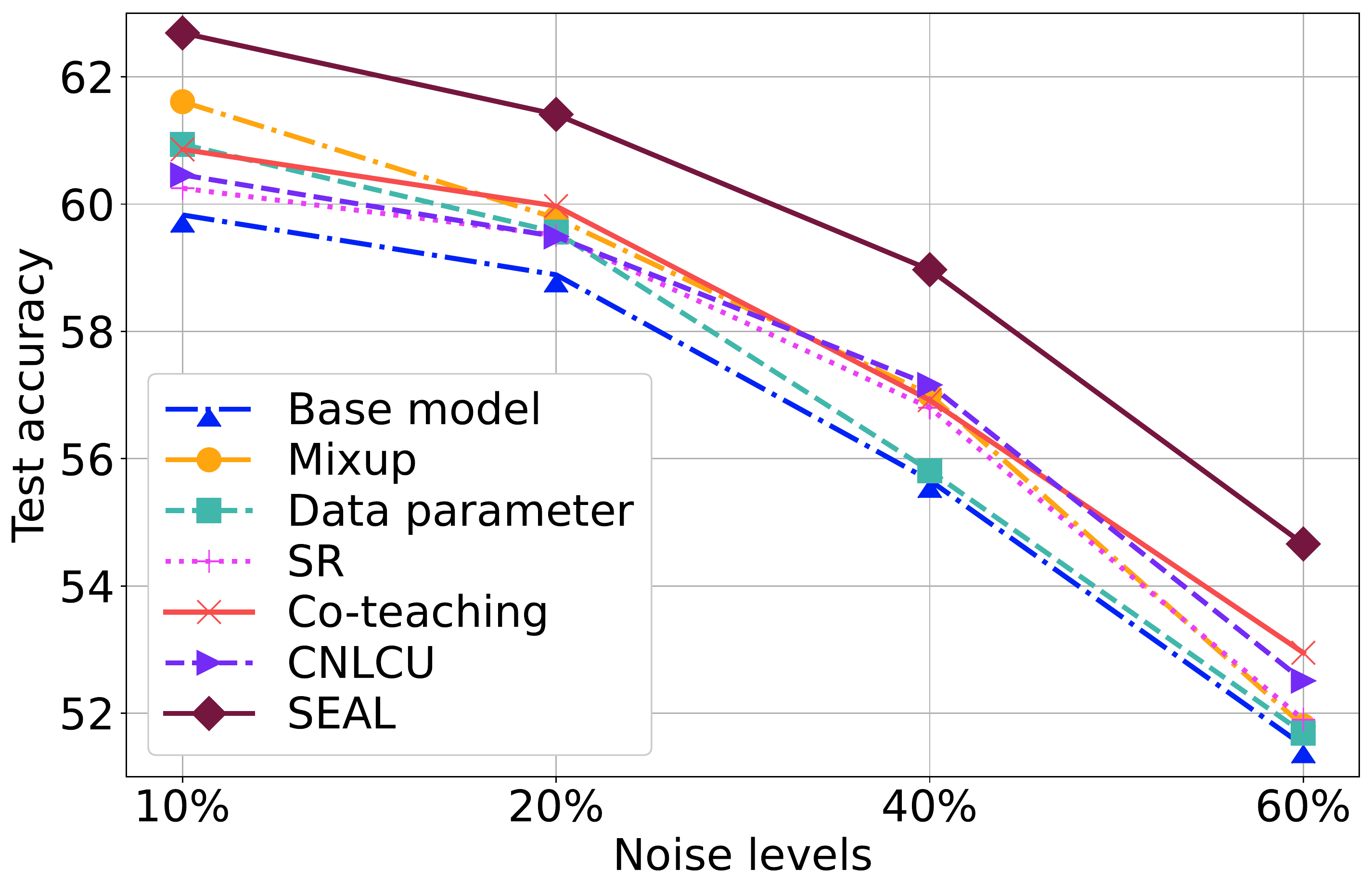}
		\label{lnl_on_syn} 
	}
	\caption{Test accuracy (Top-1) of representative LNL methods trained with controlled label noise.}
	\label{lnl} 
\end{figure*}


\begin{figure*}[t]
	\centering
	\subfigure[Real-world label noise.]{
		\includegraphics[width=0.9\columnwidth]{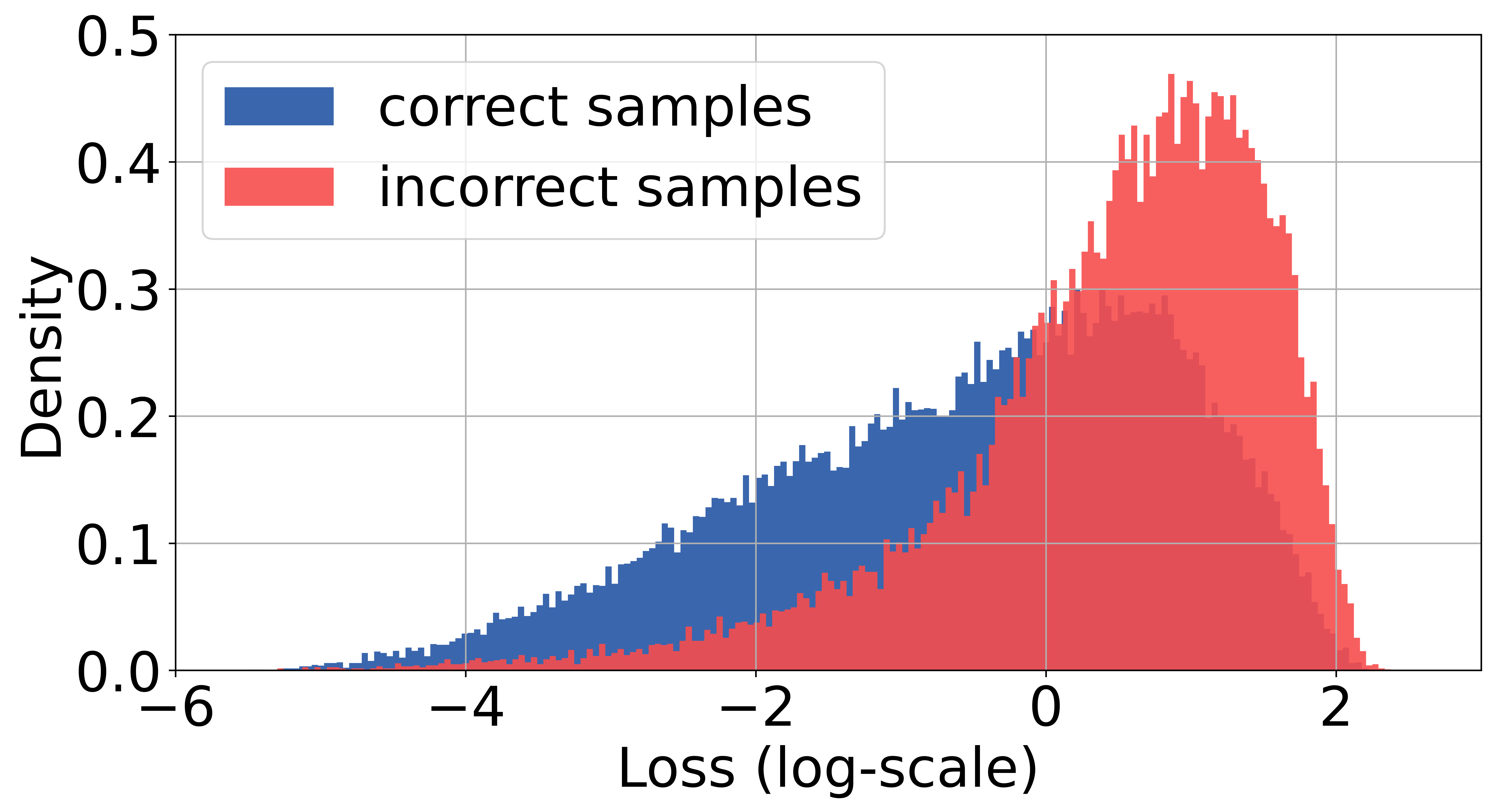}
		\label{loss_density_idn} 
	}
	\subfigure[Synthetic label noise]{
		\includegraphics[width=0.9\columnwidth]{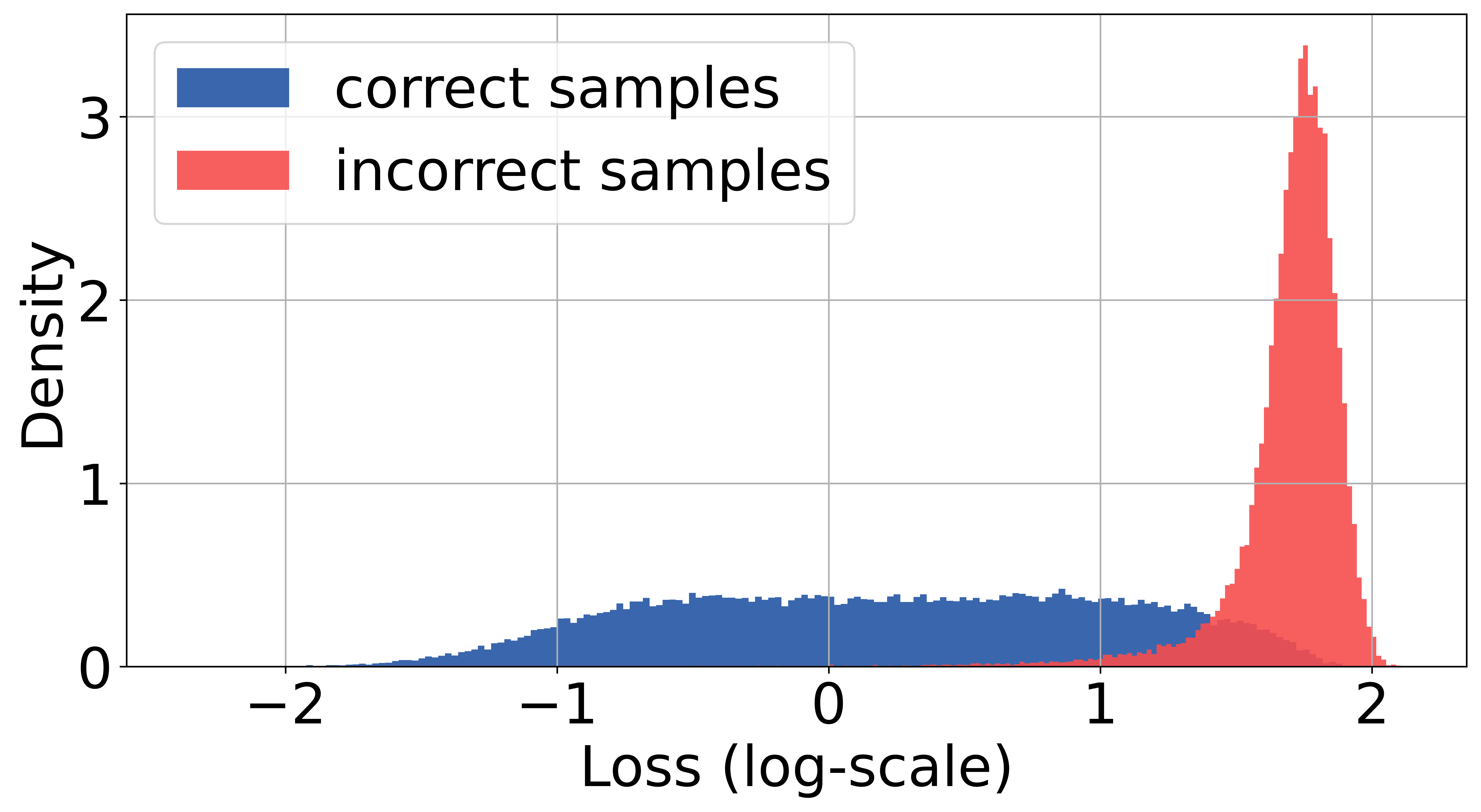}
		\label{loss_density_syn} 
	}
	\caption{Training loss distributions of correct samples and incorrect ones at the 4-th epoch with 40\% label noise.}
	\label{lnl1} 
\end{figure*}
\begin{figure*}[t]
	\centering
	\subfigure[Real-world label noise]{
		\includegraphics[width=0.9\columnwidth]{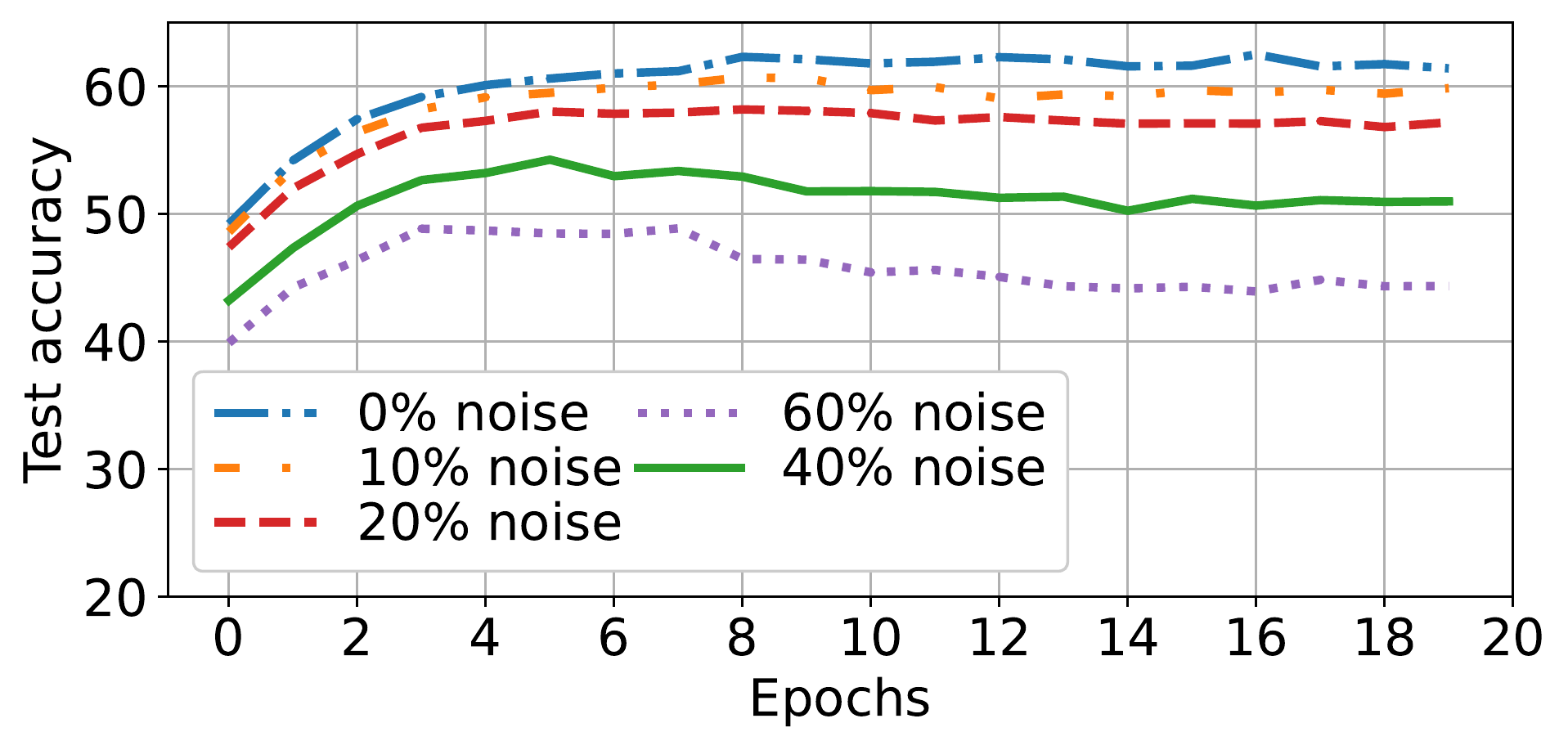}
		\label{idn_20eps} 
	}
	\subfigure[Synthetic label noise]{
		\includegraphics[width=0.9\columnwidth]{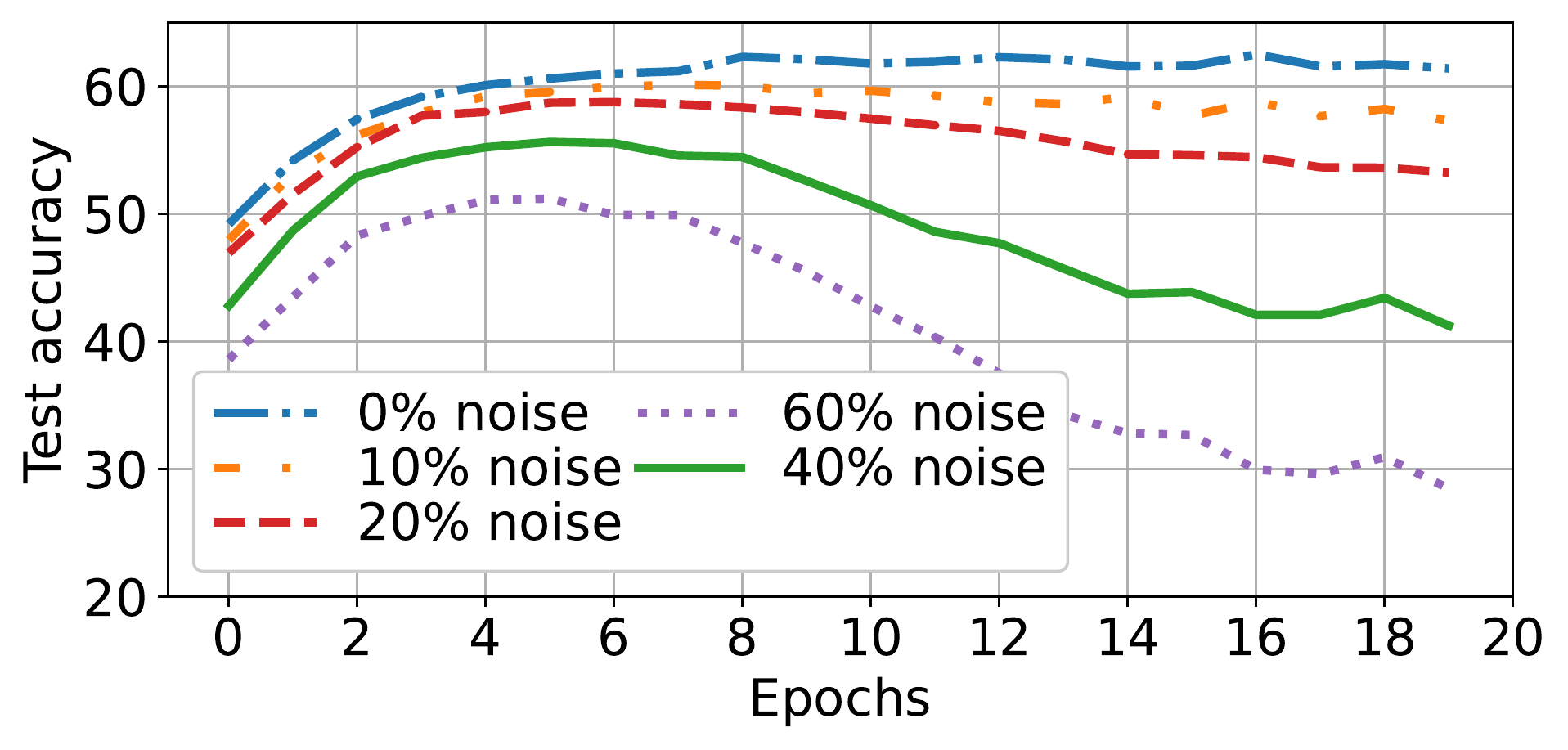}
		\label{syn_20eps} 
	}
	\caption{Test accuracy of the base model trained with controlled label noise.}
	\label{idn_syn_20eps} 
\end{figure*}
\begin{table*}[]
	\small\setlength\tabcolsep{3pt}
	\centering
	\begin{tabular}{clcc}
\toprule
\textbf{\begin{tabular}[c]{@{}c@{}}Noise\\ Sources\end{tabular}} & \multicolumn{1}{c}{\textbf{Incorrect sample}}           & \textbf{Noisy label}     & \textbf{\begin{tabular}[c]{@{}c@{}}Ground-truth label\\ (unobservable)\end{tabular}}   \\ \midrule
SC                     & (a) Rinse off the paste using warm water.      & Coloring Hair           & Making Skin Look Lighter               \\ \midrule
\multirow{2}{*}{BCC}   & (b) Mow your lawn and the leaves.                     & Lawn Care                & Cleaning up Garden            \\ \cmidrule{2-4} 
& (c) Avoid over-fertilizing your tree.                 & Growing Trees and Shrubs & Growing Fruit                 \\ \midrule
\multirow{2}{*}{TC}    & (d) Give yourself a span of time to mourn.            & Domestic Violence        & Rebuilding Life After Divorce \\ \cmidrule{2-4} 
& (e) Place the bananas on a wire rack. & Steaming Food                  & Food Preservation Techniques                      \\ \bottomrule
\end{tabular}
\caption{Five incorrect instances from three different noise sources in the NoisywikiHow dataset.}
\label{case_study}
\end{table*}

\subsection{Effectiveness of Different LNL methods}\label{exp_lnl}
\textbf{Baselines}:
We perform an extensive evaluation of the existing LNL methods on our benchmark. Seven representative baselines are involved for comparison: 
(1) \emph{\textbf{Base model}}, which finetunes the BART model with no extra LNL methods;
(2) \emph{\textbf{Mixup}}~\citep{zhang2018mixup}, which mitigates memorization of noisy labels by
DNNs regularization, i.e., introducing a data-agnostic data augmentation routine; 
(3) \emph{\textbf{Data Parameter}}~\citep{saxena2019data}, which equips learnable parameters to help DNNs generalize better via learning from easier instances first; 
(4) \emph{\textbf{SR}}~\citep{zhou2021learning}, which introduces the sparse regularization strategy, making any loss robust to noisy labels conforming to the specified assumption; 
(5) \emph{\textbf{Co-teaching}}~\citep{BoHan2018CoteachingRT}, which combats noisy labels by training two networks, and each network aims to teach the other one with clean data, i.e., the instances with small-loss;  
(6) \emph{\textbf{CNLCU}}~\citep{xia2022sample}, which considers the uncertainty of loss estimation to refine correct sample selection; 
(7) \emph{\textbf{SEAL}}~\citep{chen2021beyond}, which provides instance-dependent label correction to resist real-world noise. 
Complete experimental results and unique hyperparameters for each noise level for each baseline are in Tables~\ref{dim3} and~\ref{hyper_LNL1} in the Appendix. 

\textbf{Results}:
As Fig.~\ref{lnl_on_idn} shows, Mixup outperforms the base model with limited performance improvement. 
It is because \emph{\textbf{Mixup fails to consider the specialty of real-world label noise}} and improves generalization with a generic regularization-based method. 
The performance of Data Parameter is comparable to or slightly better than the base model under different noise levels. 
Although Data Parameter models the situation that instances within a class have different difficulty levels, it assumes \emph{small-loss training samples as correct samples and splits correct and incorrect samples via a loss-based separation}.
However, as shown in Fig.~\ref{loss_density_idn}, loss distributions of correct and incorrect data overlap closely in the real-world label noise, making \emph{\textbf{Data Parameter has no advantage under real-world label noise}}.
Similarly, Co-teaching and CNLCU fulfill sample selection following the same assumptions. They perform worse than the base model, with the exception of individual noise levels. It implies that \emph{\textbf{Co-teaching and CNLCU are inapplicable to the heterogeneous and instance-dependent label noise}}.
SR precedes the base model only at certain noise levels. This is because SR guarantees noise tolerance if and only if the label noise satisfies the instance-independent condition, which is inconsistent with noise in the real world. Hence, \emph{\textbf{the validity of SR is not ensured on our benchmark}}.
SEAL consistently outperforms the base model by a large margin on all noise levels, as SEAL provides instance-dependent label correction to combat real-world noise. However, during the correction, SEAL retrains the classifier using the averaged soft labels, introducing excessive computational overhead. 

\subsection{Real-world Noise vs. Synthetic Noise
}\label{exp_rw_vs_syn}



Aside from the real-world label noise, synthetic label noise is one of the most widely studied label noises~\citep{patrini2017making,wang2018iterative,reeve2019fast}. 
Unlike real-world noise, which is widespread in real applications, synthetic noise does not exist but is generated from artificial distributions. We further examine the differences between the two label noises.  
In this paper, synthetic label noise is implemented with symmetric label noise~\citep{han2018masking,charoenphakdee2019symmetric} (the most common synthetic noise), assuming each label has the same probability of flipping to any other class.  
We build the dataset of controlled synthetic label noise by injecting a series of synthetic label noises into clean data in a controlled manner (i.e., 10\%, 20\%, 40\%, and 60\% noise levels).  We pick the same baselines as in Section~\ref{exp_lnl}. More details are in Tables~\ref{dim5} and~\ref{hyper_LNL2} in the Appendix.


\textbf{Results}:
As shown in Fig.~\ref{lnl_on_syn}, SEAL and Mixup consistently outperform the base model, showing their advantages in combating synthetic label noise. 
Unlike the real-world label noise, 
SR is effective for the synthetic label noise and achieves improvement over the base model regardless of the noise levels since the synthetic label noise meets the instance-independent condition.  
Besides, Co-teaching, Data Parameter, and CNLCU improve the base model by an apparent margin under the synthetic label noise. In this case, as shown in Fig.~\ref{loss_density_syn}, the loss distributions of correct and incorrect samples can be well split, allowing loss-based separation to work well.
We discover that few LNL methods can effectively resist both real-world and synthetic noises simultaneously, highlighting the imperative of benchmark construction.
Many LNL approaches can mitigate the synthetic but not real-world label noise. 
It is because synthetic noise is generated from artificial distributions to approximate real-world noise. 
The mislabeled probability is independent of each instance under synthetic noise but dependent on distinct instances under real-world noise, which makes complex modeling of the latter. 
Thus, our benchmark contributes to a more systematic and comprehensive assessment of LNL methods. 
Further, since most LNL method evaluation datasets focus on the CV and ASP, our NLP benchmark facilitates the modal integrity of the existing datasets.

We also contrast the performance of the base model trained for 20 epochs under real-world label noise and synthetic label noise. In Fig.~\ref{idn_syn_20eps}, as the running epochs and noise levels increase, the test accuracy curve with the real-world noise (Fig.~\ref{idn_20eps}) is much flatter than that with the synthetic noise (Fig.~\ref{syn_20eps}) at the same noise level (e.g., with 40\% and 60\% noise). It demonstrates that the model generalizes much better under real-world noise than synthetic noise of the same noise level.

\subsection{Case Study}\label{exp_quality}
We construct a benchmark encompassing real-world noise involving multiple noise sources with minimal human supervision, which is analogous to human errors during annotation. 
To observe the dataset more clearly and intuitively, we randomly select five incorrect instances (i.e., samples with noisy labels) across multiple noise sources. As indicated in Table~\ref{case_study}, we find it difficult to determine whether the sample contains noise. On the other hand, for any sample, the noise label and the respective ground-truth label are overly similar, making it challenging to distinguish one from another. 

\section{Conclusion}
\label{sec:bibtex}

In this paper, we study the problem of learning with noisy labels and establish an NLP benchmark called NoisywikiHow with minimal human supervision, which contains more than 89K procedural events with heterogeneous and controlled real-world label noise.
Experimental results reveal several new findings. 
(1) Some widely accepted LNL methods are not always impactful, especially with real-world label noise. 
(2) Different noise sources may have varying difficulties resisting label noise, although they are all from real-world noise. 
(3) Few LNL methods can effectively combat real-world noise and synthetic noise at the same time. 
(4) The model trained under the real-world label noise has better generalization performance. 
%

\section*{Limitations}
In this paper, we simplify intention identification into a sentence classification task, i.e., exploiting a specific procedural event in an event process to predict the intention of the whole event process. A more realistic way to model this task is to enter the entire event process rather than a single event. We will go into more detail about this type of task in future work.
%

\section*{Ethics Statement}
This work presents NoisywikiHow, a free and open dataset for the research community to study learning with noisy labels. Since the data in NoisywikiHow is constructed based on the wikiHow website, which is free and open for academic usage, there is no privacy issue. 
We declare that all information in this paper has been obtained and presented following the \href{https://www.aclweb.org/portal/content/acl-code-ethics}{ACL Ethics Policy}. As required by these rules and conduct, we have fully cited and referenced all material and results that are not original to this work.

\section*{Acknowledgements}
We thank the anonymous reviewers for their constructive comments, and gratefully acknowledge the support of the Technological Innovation “2030 Megaproject” - New Generation Artificial Intelligence of China (2020AAA0106501), and the National Natural Science Foundation of China (U22B2059, 62176079, 62106061). 

\bibliography{anthology,custom}

\begin{thebibliography}{62}
\expandafter\ifx\csname natexlab\endcsname\relax\def\natexlab#1{#1}\fi

\bibitem[{Beigman and Klebanov(2009)}]{beigman2009learning}
Eyal Beigman and Beata~Beigman Klebanov. 2009.
\newblock Learning with annotation noise.
\newblock In \emph{Proceedings of the Joint Conference of the 47th Annual
  Meeting of the ACL and the 4th International Joint Conference on Natural
  Language Processing of the AFNLP}, pages 280--287.

\bibitem[{Bergstra et~al.(2013)Bergstra, Yamins, and Cox}]{bergstra2013making}
James Bergstra, Daniel Yamins, and David Cox. 2013.
\newblock Making a science of model search: Hyperparameter optimization in
  hundreds of dimensions for vision architectures.
\newblock In \emph{International conference on machine learning}, pages
  115--123. PMLR.

\bibitem[{Burghardt et~al.(2018)Burghardt, Hogg, and
  Lerman}]{burghardt2018quantifying}
Keith Burghardt, Tad Hogg, and Kristina Lerman. 2018.
\newblock Quantifying the impact of cognitive biases in question-answering
  systems.
\newblock In \emph{Twelfth International AAAI Conference on Web and Social
  Media}.

\bibitem[{Cao et~al.(2019)Cao, Wei, Gaidon, Arechiga, and Ma}]{cao2019learning}
Kaidi Cao, Colin Wei, Adrien Gaidon, Nikos Arechiga, and Tengyu Ma. 2019.
\newblock Learning imbalanced datasets with label-distribution-aware margin
  loss.
\newblock \emph{Advances in neural information processing systems}, 32.

\bibitem[{Charoenphakdee et~al.(2019)Charoenphakdee, Lee, and
  Sugiyama}]{charoenphakdee2019symmetric}
Nontawat Charoenphakdee, Jongyeong Lee, and Masashi Sugiyama. 2019.
\newblock On symmetric losses for learning from corrupted labels.
\newblock In \emph{International Conference on Machine Learning}, pages
  961--970. PMLR.

\bibitem[{Chen et~al.(2020)Chen, Zhang, Wang, and Roth}]{chen2020you}
Muhao Chen, Hongming Zhang, Haoyu Wang, and Dan Roth. 2020.
\newblock What are you trying to do? semantic typing of event processes.
\newblock In \emph{Proceedings of the 24th Conference on Computational Natural
  Language Learning}, pages 531--542.

\bibitem[{Chen et~al.(2021)Chen, Ye, Chen, Zhao, and Heng}]{chen2021beyond}
Pengfei Chen, Junjie Ye, Guangyong Chen, Jingwei Zhao, and Pheng-Ann Heng.
  2021.
\newblock Beyond class-conditional assumption: A primary attempt to combat
  instance-dependent label noise.
\newblock In \emph{Proceedings of the AAAI Conference on Artificial
  Intelligence}, volume~35, pages 11442--11450.

\bibitem[{Devlin et~al.(2019)Devlin, Chang, Lee, and
  Toutanova}]{JacobDevlin2018BERTPO}
Jacob Devlin, Ming-Wei Chang, Kenton Lee, and Kristina Toutanova. 2019.
\newblock Bert: Pre-training of deep bidirectional transformers for language
  understanding.
\newblock In \emph{North American Chapter of the Association for Computational
  Linguistics}.

\bibitem[{Fleiss(1971)}]{fleiss1971measuring}
Joseph~L Fleiss. 1971.
\newblock Measuring nominal scale agreement among many raters.
\newblock \emph{Psychological bulletin}, 76(5):378.

\bibitem[{Fonseca et~al.(2019{\natexlab{a}})Fonseca, Plakal, Ellis, Font,
  Favory, and Serra}]{fonseca2019learning}
Eduardo Fonseca, Manoj Plakal, Daniel~PW Ellis, Frederic Font, Xavier Favory,
  and Xavier Serra. 2019{\natexlab{a}}.
\newblock Learning sound event classifiers from web audio with noisy labels.
\newblock In \emph{ICASSP 2019-2019 IEEE International Conference on Acoustics,
  Speech and Signal Processing (ICASSP)}, pages 21--25. IEEE.

\bibitem[{Fonseca et~al.(2019{\natexlab{b}})Fonseca, Plakal, Font, Ellis, and
  Serra}]{fonseca2019audio}
Eduardo Fonseca, Manoj Plakal, Frederic Font, Daniel~PW Ellis, and Xavier
  Serra. 2019{\natexlab{b}}.
\newblock Audio tagging with noisy labels and minimal supervision.
\newblock \emph{arXiv preprint arXiv:1906.02975}.

\bibitem[{Gemmeke et~al.(2017)Gemmeke, Ellis, Freedman, Jansen, Lawrence,
  Moore, Plakal, and Ritter}]{gemmeke2017audio}
Jort~F Gemmeke, Daniel~PW Ellis, Dylan Freedman, Aren Jansen, Wade Lawrence,
  R~Channing Moore, Manoj Plakal, and Marvin Ritter. 2017.
\newblock Audio set: An ontology and human-labeled dataset for audio events.
\newblock In \emph{2017 IEEE international conference on acoustics, speech and
  signal processing (ICASSP)}, pages 776--780. IEEE.

\bibitem[{Han et~al.(2018{\natexlab{a}})Han, Yao, Niu, Zhou, Tsang, Zhang, and
  Sugiyama}]{han2018masking}
Bo~Han, Jiangchao Yao, Gang Niu, Mingyuan Zhou, Ivor Tsang, Ya~Zhang, and
  Masashi Sugiyama. 2018{\natexlab{a}}.
\newblock Masking: A new perspective of noisy supervision.
\newblock \emph{Advances in neural information processing systems}, 31.

\bibitem[{Han et~al.(2021)Han, Yao, Liu, Niu, Tsang, Kwok, and
  Sugiyama}]{han2020survey}
Bo~Han, Quanming Yao, Tongliang Liu, Gang Niu, Ivor~W Tsang, James~T Kwok, and
  Masashi Sugiyama. 2021.
\newblock A survey of label-noise representation learning: Past, present and
  future.

\bibitem[{Han et~al.(2018{\natexlab{b}})Han, Yao, Yu, Niu, Xu, Hu, Tsang, and
  Sugiyama}]{BoHan2018CoteachingRT}
Bo~Han, Quanming Yao, Xingrui Yu, Gang Niu, Miao Xu, Weihua Hu, Ivor~W. Tsang,
  and Masashi Sugiyama. 2018{\natexlab{b}}.
\newblock Co-teaching: Robust training of deep neural networks with extremely
  noisy labels.
\newblock In \emph{Neural Information Processing Systems}.

\bibitem[{Hedderich et~al.(2021)Hedderich, Zhu, and
  Klakow}]{hedderich2021analysing}
Michael~A Hedderich, Dawei Zhu, and Dietrich Klakow. 2021.
\newblock Analysing the noise model error for realistic noisy label data.
\newblock In \emph{Proceedings of the AAAI Conference on Artificial
  Intelligence}, volume~35, pages 7675--7684.

\bibitem[{Hollenstein et~al.(2016)Hollenstein, Schneider, and
  Webber}]{hollenstein2016inconsistency}
Nora Hollenstein, Nathan Schneider, and Bonnie Webber. 2016.
\newblock Inconsistency detection in semantic annotation.
\newblock In \emph{Proceedings of the Tenth International Conference on
  Language Resources and Evaluation (LREC'16)}, pages 3986--3990.

\bibitem[{Honnibal and Montani(2017)}]{honnibal2017spacy}
Matthew Honnibal and Ines Montani. 2017.
\newblock spacy 2: Natural language understanding with bloom embeddings,
  convolutional neural networks and incremental parsing.
\newblock \emph{To appear}, 7(1):411--420.

\bibitem[{Huang et~al.(2016)Huang, Cassidy, Feng, Ji, Voss, Han, and
  Sil}]{huang2016liberal}
Lifu Huang, Taylor Cassidy, Xiaocheng Feng, Heng Ji, Clare Voss, Jiawei Han,
  and Avirup Sil. 2016.
\newblock Liberal event extraction and event schema induction.
\newblock In \emph{Proceedings of the 54th Annual Meeting of the Association
  for Computational Linguistics (Volume 1: Long Papers)}, pages 258--268.

\bibitem[{Jiang et~al.(2020)Jiang, Huang, Liu, and Yang}]{jiang2020beyond}
Lu~Jiang, Di~Huang, Mason Liu, and Weilong Yang. 2020.
\newblock Beyond synthetic noise: Deep learning on controlled noisy labels.
\newblock In \emph{International Conference on Machine Learning}, pages
  4804--4815. PMLR.

\bibitem[{Kang et~al.(2020)Kang, Xie, Rohrbach, Yan, Gordo, Feng, and
  Kalantidis}]{kang2019decoupling}
Bingyi Kang, Saining Xie, Marcus Rohrbach, Zhicheng Yan, Albert Gordo, Jiashi
  Feng, and Yannis Kalantidis. 2020.
\newblock Decoupling representation and classifier for long-tailed recognition.
\newblock In \emph{Eighth International Conference on Learning Representations
  (ICLR)}.

\bibitem[{Klebanov and Beigman(2009)}]{klebanov2009squibs}
Beata~Beigman Klebanov and Eyal Beigman. 2009.
\newblock Squibs: From annotator agreement to noise models.
\newblock \emph{Computational Linguistics}, 35(4):495--503.

\bibitem[{Klebanov et~al.(2008)Klebanov, Beigman, and
  Diermeier}]{klebanov2008analyzing}
Beata~Beigman Klebanov, Eyal Beigman, and Daniel Diermeier. 2008.
\newblock Analyzing disagreements.
\newblock In \emph{Coling 2008: Proceedings of the workshop on Human Judgements
  in Computational Linguistics}, pages 2--7.

\bibitem[{Kouzis-Loukas(2016)}]{kouzis2016learning}
Dimitrios Kouzis-Loukas. 2016.
\newblock \emph{Learning Scrapy}.
\newblock Packt Publishing Ltd.

\bibitem[{Lan et~al.(2020)Lan, Chen, Goodman, Gimpel, Sharma, and
  Soricut}]{lan2020albert}
Zhenzhong Lan, Mingda Chen, Sebastian Goodman, Kevin Gimpel, Piyush Sharma, and
  Radu Soricut. 2020.
\newblock Albert: A lite bert for self-supervised learning of language
  representations.
\newblock In \emph{International Conference on Learning Representations}.

\bibitem[{Lee et~al.(2018)Lee, He, Zhang, and Yang}]{lee2018cleannet}
Kuang-Huei Lee, Xiaodong He, Lei Zhang, and Linjun Yang. 2018.
\newblock Cleannet: Transfer learning for scalable image classifier training
  with label noise.
\newblock In \emph{Proceedings of the IEEE conference on computer vision and
  pattern recognition}, pages 5447--5456.

\bibitem[{Lewis et~al.(2020)Lewis, Liu, Goyal, Ghazvininejad, Mohamed, Levy,
  Stoyanov, and Zettlemoyer}]{lewis2020bart}
Mike Lewis, Yinhan Liu, Naman Goyal, Marjan Ghazvininejad, Abdelrahman Mohamed,
  Omer Levy, Veselin Stoyanov, and Luke Zettlemoyer. 2020.
\newblock Bart: Denoising sequence-to-sequence pre-training for natural
  language generation, translation, and comprehension.
\newblock In \emph{Proceedings of the 58th Annual Meeting of the Association
  for Computational Linguistics}, pages 7871--7880.

\bibitem[{Li et~al.(2017)Li, Wang, Li, Agustsson, and
  Van~Gool}]{li2017webvision}
Wen Li, Limin Wang, Wei Li, Eirikur Agustsson, and Luc Van~Gool. 2017.
\newblock Webvision database: Visual learning and understanding from web data.
\newblock \emph{arXiv preprint arXiv:1708.02862}.

\bibitem[{Liu and Singh(2004)}]{liu2004conceptnet}
Hugo Liu and Push Singh. 2004.
\newblock Conceptnet—a practical commonsense reasoning tool-kit.
\newblock \emph{BT technology journal}, 22(4):211--226.

\bibitem[{Liu et~al.(2019{\natexlab{a}})Liu, Ott, Goyal, Du, Joshi, Chen, Levy,
  Lewis, Zettlemoyer, and Stoyanov}]{YinhanLiu2019RoBERTaAR}
Yinhan Liu, Myle Ott, Naman Goyal, Jingfei Du, Mandar Joshi, Danqi Chen, Omer
  Levy, Mike Lewis, Luke Zettlemoyer, and Veselin Stoyanov. 2019{\natexlab{a}}.
\newblock Roberta: A robustly optimized bert pretraining approach.
\newblock \emph{arXiv: Computation and Language}.

\bibitem[{Liu et~al.(2019{\natexlab{b}})Liu, Miao, Zhan, Wang, Gong, and
  Yu}]{liu2019large}
Ziwei Liu, Zhongqi Miao, Xiaohang Zhan, Jiayun Wang, Boqing Gong, and Stella~X
  Yu. 2019{\natexlab{b}}.
\newblock Large-scale long-tailed recognition in an open world.
\newblock In \emph{Proceedings of the IEEE/CVF Conference on Computer Vision
  and Pattern Recognition}, pages 2537--2546.

\bibitem[{Lukasik et~al.(2020)Lukasik, Bhojanapalli, Menon, and
  Kumar}]{lukasik2020does}
Michal Lukasik, Srinadh Bhojanapalli, Aditya Menon, and Sanjiv Kumar. 2020.
\newblock Does label smoothing mitigate label noise?
\newblock In \emph{International Conference on Machine Learning}, pages
  6448--6458. PMLR.

\bibitem[{Min et~al.(2020)Min, Chan, and Zhao}]{min2020towards}
Bonan Min, Yee~Seng Chan, and Lingjun Zhao. 2020.
\newblock Towards few-shot event mention retrieval: An evaluation framework and
  a siamese network approach.
\newblock In \emph{Proceedings of the 12th Language Resources and Evaluation
  Conference}, pages 1747--1752.

\bibitem[{Northcutt et~al.(2021)Northcutt, Athalye, and
  Mueller}]{northcutt2021pervasive}
Curtis~G Northcutt, Anish Athalye, and Jonas Mueller. 2021.
\newblock Pervasive label errors in test sets destabilize machine learning
  benchmarks.
\newblock In \emph{Thirty-fifth Conference on Neural Information Processing
  Systems Datasets and Benchmarks Track (Round 1)}.

\bibitem[{Patrini et~al.(2017)Patrini, Rozza, Krishna~Menon, Nock, and
  Qu}]{patrini2017making}
Giorgio Patrini, Alessandro Rozza, Aditya Krishna~Menon, Richard Nock, and
  Lizhen Qu. 2017.
\newblock Making deep neural networks robust to label noise: A loss correction
  approach.
\newblock In \emph{Proceedings of the IEEE conference on computer vision and
  pattern recognition}, pages 1944--1952.

\bibitem[{Pepe et~al.(2022)Pepe, Barba, Blloshmi, and Navigli}]{pepe2022steps}
Sveva Pepe, Edoardo Barba, Rexhina Blloshmi, and Roberto Navigli. 2022.
\newblock Steps: Semantic typing of event processes with a sequence-to-sequence
  approach.

\bibitem[{Radford et~al.(2019)Radford, Wu, Child, Luan, Amodei, and
  Sutskever}]{radford2019language}
Alec Radford, Jeffrey Wu, Rewon Child, David Luan, Dario Amodei, and Ilya
  Sutskever. 2019.
\newblock Language models are unsupervised multitask learners.
\newblock \emph{OpenAI blog}, 1(8):9.

\bibitem[{Raffel et~al.(2020)Raffel, Shazeer, Roberts, Lee, Narang, Matena,
  Zhou, Li, and Liu}]{raffel2020exploring}
Colin Raffel, Noam Shazeer, Adam Roberts, Katherine Lee, Sharan Narang, Michael
  Matena, Yanqi Zhou, Wei Li, and Peter~J Liu. 2020.
\newblock Exploring the limits of transfer learning with a unified text-to-text
  transformer.
\newblock \emph{Journal of Machine Learning Research}, 21:1--67.

\bibitem[{Reeve and Kab{\'a}n(2019)}]{reeve2019fast}
Henry Reeve and Ata Kab{\'a}n. 2019.
\newblock Fast rates for a knn classifier robust to unknown asymmetric label
  noise.
\newblock In \emph{International Conference on Machine Learning}, pages
  5401--5409. PMLR.

\bibitem[{Reidsma and op~den Akker(2008)}]{reidsma2008exploiting}
Dennis Reidsma and Rieks op~den Akker. 2008.
\newblock Exploiting ‘subjective’annotations.
\newblock In \emph{Coling 2008: Proceedings of the workshop on Human Judgements
  in Computational Linguistics}, pages 8--16.

\bibitem[{Russakovsky et~al.(2015)Russakovsky, Deng, Su, Krause, Satheesh, Ma,
  Huang, Karpathy, Khosla, Bernstein et~al.}]{russakovsky2015imagenet}
Olga Russakovsky, Jia Deng, Hao Su, Jonathan Krause, Sanjeev Satheesh, Sean Ma,
  Zhiheng Huang, Andrej Karpathy, Aditya Khosla, Michael Bernstein, et~al.
  2015.
\newblock Imagenet large scale visual recognition challenge.
\newblock \emph{International Journal of Computer Vision}, 115(3):211--252.

\bibitem[{Sap et~al.(2019)Sap, Le~Bras, Allaway, Bhagavatula, Lourie, Rashkin,
  Roof, Smith, and Choi}]{sap2019atomic}
Maarten Sap, Ronan Le~Bras, Emily Allaway, Chandra Bhagavatula, Nicholas
  Lourie, Hannah Rashkin, Brendan Roof, Noah~A Smith, and Yejin Choi. 2019.
\newblock Atomic: An atlas of machine commonsense for if-then reasoning.
\newblock In \emph{Proceedings of the AAAI Conference on Artificial
  Intelligence}, volume~33, pages 3027--3035.

\bibitem[{Saxena et~al.(2019)Saxena, Tuzel, and DeCoste}]{saxena2019data}
Shreyas Saxena, Oncel Tuzel, and Dennis DeCoste. 2019.
\newblock Data parameters: A new family of parameters for learning a
  differentiable curriculum.
\newblock \emph{Advances in Neural Information Processing Systems}, 32.

\bibitem[{Song et~al.(2019)Song, Kim, and Lee}]{song2019selfie}
Hwanjun Song, Minseok Kim, and Jae-Gil Lee. 2019.
\newblock Selfie: Refurbishing unclean samples for robust deep learning.
\newblock In \emph{International Conference on Machine Learning}, pages
  5907--5915. PMLR.

\bibitem[{Song et~al.(2022)Song, Kim, Park, Shin, and Lee}]{song2022learning}
Hwanjun Song, Minseok Kim, Dongmin Park, Yooju Shin, and Jae-Gil Lee. 2022.
\newblock Learning from noisy labels with deep neural networks: A survey.
\newblock \emph{IEEE Transactions on Neural Networks and Learning Systems}.

\bibitem[{Sutskever et~al.(2014)Sutskever, Vinyals, and
  Le}]{sutskever2014sequence}
Ilya Sutskever, Oriol Vinyals, and Quoc~V Le. 2014.
\newblock Sequence to sequence learning with neural networks.
\newblock \emph{Advances in neural information processing systems}, 27.

\bibitem[{Van~Horn et~al.(2018)Van~Horn, Mac~Aodha, Song, Cui, Sun, Shepard,
  Adam, Perona, and Belongie}]{van2018inaturalist}
Grant Van~Horn, Oisin Mac~Aodha, Yang Song, Yin Cui, Chen Sun, Alex Shepard,
  Hartwig Adam, Pietro Perona, and Serge Belongie. 2018.
\newblock The inaturalist species classification and detection dataset.
\newblock In \emph{Proceedings of the IEEE conference on computer vision and
  pattern recognition}, pages 8769--8778.

\bibitem[{Wang et~al.(2019)Wang, Singh, Michael, Hill, Levy, and
  Bowman}]{wang2018glue}
Alex Wang, Amanpreet Singh, Julian Michael, Felix Hill, Omer Levy, and Samuel~R
  Bowman. 2019.
\newblock Glue: A multi-task benchmark and analysis platform for natural
  language understanding.
\newblock In \emph{International Conference on Learning Representations}.

\bibitem[{Wang et~al.(2018)Wang, Liu, Ma, Bailey, Zha, Song, and
  Xia}]{wang2018iterative}
Yisen Wang, Weiyang Liu, Xingjun Ma, James Bailey, Hongyuan Zha, Le~Song, and
  Shu-Tao Xia. 2018.
\newblock Iterative learning with open-set noisy labels.
\newblock In \emph{Proceedings of the IEEE conference on computer vision and
  pattern recognition}, pages 8688--8696.

\bibitem[{Wu et~al.(2022{\natexlab{a}})Wu, Ding, Tang, Zhang, Qin, and
  Liu}]{wu2022stgn}
Tingting Wu, Xiao Ding, Minji Tang, Hao Zhang, Bing Qin, and Ting Liu.
  2022{\natexlab{a}}.
\newblock Stgn: an implicit regularization method for learning with noisy
  labels in natural language processing.
\newblock In \emph{Proceedings of the 2022 Conference on Empirical Methods in
  Natural Language Processing}, page 7587–7598.

\bibitem[{Wu et~al.(2022{\natexlab{b}})Wu, Ding, Zhang, Gao, Du, Qin, and
  Liu}]{wu2022discrimloss}
Tingting Wu, Xiao Ding, Hao Zhang, Jinglong Gao, Li~Du, Bing Qin, and Ting Liu.
  2022{\natexlab{b}}.
\newblock Discrimloss: A universal loss for hard samples and incorrect samples
  discrimination.
\newblock \emph{arXiv preprint arXiv:2208.09884}.

\bibitem[{Xia et~al.(2022)Xia, Liu, Han, Gong, Yu, Niu, and
  Sugiyama}]{xia2022sample}
Xiaobo Xia, Tongliang Liu, Bo~Han, Mingming Gong, Jun Yu, Gang Niu, and Masashi
  Sugiyama. 2022.
\newblock Sample selection with uncertainty of losses for learning with noisy
  labels.
\newblock In \emph{International Conference on Learning Representations}.

\bibitem[{Xiao et~al.(2015)Xiao, Xia, Yang, Huang, and Wang}]{xiao2015learning}
Tong Xiao, Tian Xia, Yi~Yang, Chang Huang, and Xiaogang Wang. 2015.
\newblock Learning from massive noisy labeled data for image classification.
\newblock In \emph{Proceedings of the IEEE conference on computer vision and
  pattern recognition}, pages 2691--2699.

\bibitem[{Yang et~al.(2019)Yang, Dai, Yang, Carbonell, Salakhutdinov, and
  Le}]{ZhilinYang2019XLNetGA}
Zhilin Yang, Zihang Dai, Yiming Yang, Jaime~G. Carbonell, Ruslan Salakhutdinov,
  and Quoc~V. Le. 2019.
\newblock Xlnet: Generalized autoregressive pretraining for language
  understanding.
\newblock In \emph{Neural Information Processing Systems}.

\bibitem[{Yu et~al.(2021)Yu, Yin, Gupta, and Roth}]{yu2021event}
Xiaodong Yu, Wenpeng Yin, Nitish Gupta, and Dan Roth. 2021.
\newblock Event linking: Grounding event mentions to wikipedia.
\newblock \emph{arXiv preprint arXiv:2112.07888}.

\bibitem[{Zhang et~al.(2017{\natexlab{a}})Zhang, Bengio, Hardt, Recht, and
  Vinyals}]{ChiyuanZhang2017UnderstandingDL}
Chiyuan Zhang, Samy Bengio, Moritz Hardt, Benjamin Recht, and Oriol Vinyals.
  2017{\natexlab{a}}.
\newblock Understanding deep learning requires rethinking generalization.
\newblock In \emph{International Conference on Learning Representations}.

\bibitem[{Zhang et~al.(2018)Zhang, Cisse, Dauphin, and
  Lopez-Paz}]{zhang2018mixup}
Hongyi Zhang, Moustapha Cisse, Yann~N Dauphin, and David Lopez-Paz. 2018.
\newblock mixup: Beyond empirical risk minimization.
\newblock In \emph{International Conference on Learning Representations}.

\bibitem[{Zhang et~al.(2020{\natexlab{a}})Zhang, Lyu, and
  Callison-Burch}]{zhang2020intent}
Li~Zhang, Qing Lyu, and Chris Callison-Burch. 2020{\natexlab{a}}.
\newblock Intent detection with wikihow.
\newblock In \emph{Proceedings of the 1st Conference of the Asia-Pacific
  Chapter of the Association for Computational Linguistics and the 10th
  International Joint Conference on Natural Language Processing}, pages
  328--333.

\bibitem[{Zhang et~al.(2020{\natexlab{b}})Zhang, Lyu, and
  Callison-Burch}]{zhang2020reasoning}
Li~Zhang, Qing Lyu, and Chris Callison-Burch. 2020{\natexlab{b}}.
\newblock Reasoning about goals, steps, and temporal ordering with wikihow.
\newblock In \emph{Proceedings of the 2020 Conference on Empirical Methods in
  Natural Language Processing (EMNLP)}, pages 4630--4639.

\bibitem[{Zhang et~al.(2017{\natexlab{b}})Zhang, Zhong, Chen, Angeli, and
  Manning}]{zhang2017position}
Yuhao Zhang, Victor Zhong, Danqi Chen, Gabor Angeli, and Christopher~D Manning.
  2017{\natexlab{b}}.
\newblock Position-aware attention and supervised data improve slot filling.
\newblock In \emph{Conference on Empirical Methods in Natural Language
  Processing}.

\bibitem[{Zhou et~al.(2020)Zhou, Cui, Wei, and Chen}]{zhou2020bbn}
Boyan Zhou, Quan Cui, Xiu-Shen Wei, and Zhao-Min Chen. 2020.
\newblock Bbn: Bilateral-branch network with cumulative learning for
  long-tailed visual recognition.
\newblock In \emph{Proceedings of the IEEE/CVF conference on computer vision
  and pattern recognition}, pages 9719--9728.

\bibitem[{Zhou et~al.(2021)Zhou, Liu, Wang, Zhai, Jiang, and
  Ji}]{zhou2021learning}
Xiong Zhou, Xianming Liu, Chenyang Wang, Deming Zhai, Junjun Jiang, and
  Xiangyang Ji. 2021.
\newblock Learning with noisy labels via sparse regularization.
\newblock In \emph{Proceedings of the IEEE/CVF International Conference on
  Computer Vision}, pages 72--81.

\end{thebibliography}
\bibliographystyle{acl_natbib}

\appendix

\section{Details of Dataset Construction}
\label{app:data_construction}
\subsection{Crawling Strategy}\label{app:crawling}
According to wikiHow's crawler rules,\footnote{\url{https://www.wikihow.com/robots.txt}} we use the crawling platform Scrapy~\citep{kouzis2016learning} to crawl all the articles in the 19 top-level categories (e.g., \emph{Arts and Entertainment}, \emph{Computers and Electronics}, etc.) of the latest wikiHow website, 
with a total of 100,623 pages (how-to articles), including 1,407,306 samples in 3,334 categories, as shown in Table~\ref{filter_statistic}.

\subsection{Filtering Strategies}\label{app:filtering}
In the main paper, we apply a collection of filters to ensure low-quality instances removal, better dataset division, and task effectiveness. The details of each filter are as follows:

\textbf{Sample Length Filter}:
We remove instances with overly short or long event descriptions or with icon information. As too-short events may be less informative, too-long depictions may exceed the length restriction of the pre-trained language model. Icons in events present rich text starting with ``smallUrl'' without specific semantic information and may interfere with the understanding of procedural events.

\textbf{Format Normalization}:
We observe that some identical event descriptions would be slightly different in distinct articles (e.g., ``\emph{Click Defragment \textbf{Y}our \textbf{H}ard \textbf{D}rive.}'' and ``\emph{Click Defragment \textbf{y}our \textbf{h}ard \textbf{d}rive.}''). 
Prior to the deduplication procedure, we devise format standardization operations. 
The manipulations involve standardizing varied languages and symbols with Unidecode, stopword exclusion and lemmatization with spaCy~\citep{honnibal2017spacy}, word segmentation \& POS tagging by applying the model ``en\_core\_web\_sm'' in spaCy and reserving events containing verbs.

\textbf{Deduplication}:
We first apply inter-class deduplication to remove instances with labels of multiple categories. Then, we filter out repeated samples to achieve in-class deduplication. After the deduplication operation, each procedural event (i.e., event) corresponds to a unique event intent (i.e., category).

\textbf{TF-IDF Filter}:
We exploit the TF-IDF filter to preclude events from being overly uninformative when identifying the corresponding event intent and guarantee the instances are representative. Specifically, each wikiHow article is considered a document. We calculate the TF-IDF for each token and preserve only the events containing keywords. 
In this context, \emph{keywords} refer to tokens whose TF-IDF values are in the top 10\% in decreasing order. Each article includes a minimum of 3 and a maximum of 10 keywords.


\textbf{Label Filter}:
We filter labels by manual annotation to retain categories depicting only events. For human labeling, we used three graduate students from the NLP field. They were educated for two hours about annotation strategy before the labeling process. 
Specifically, we use~\citet{min2020towards}'s and~\citet{yu2021event}'s definitions of \emph{event mention} (i.e., an event with surrounding context (text)) as guidelines for annotating events. 
In addition, categories exhibit a hierarchical structure. Typically, the descriptions of the upper categories are relatively general and vague (e.g., \emph{Cleaning}), while the more fine-grained categories have more specific intentions (e.g., \emph{Kitchen Cleaning}, \emph{Cleaning Metals}).
Accordingly, we label the category with the finest granularity as an event except for two cases. 
\begin{itemize}[itemsep=0pt, topsep=3pt, partopsep=3pt]
\item If a candidate category has a broad intent meaning (e.g., \emph{\textbf{Selling}} in \emph{Finance and Business \(\gg\) Managing Your Money \(\gg\) Making Money \(\gg\) Selling}), it will not be considered an event. 
\item If it is difficult to distinguish semantically between two candidate categories, the category with the larger sample size is designated as an event. 
For example, in hierarchical categories (\emph{Hobbies and Crafts \(\gg\) Crafts \(\gg\) Needlework \(\gg\) Knitting and Crochet \(\gg\) Crochet \(\gg\) Crochet Stitches}), we label \emph{\textbf{Crochet}} (with 1,263 samples) as an event rather than \emph{\textbf{Crocheet Stitches}} (with 445 samples).
\end{itemize} 
This annotation strategy facilitates the balance between definite event intent and sample size.
Sample size and class info reserved after data cleaning are provided in Table~\ref{filter_statistic}.

\subsection{Mapping from Noise Sources to Classes
}\label{app:data_analysis}
In the main paper, we briefly present the correspondence between noise sources and task categories. In particular, we first define 54 tail categories, each containing no more than 400 samples.  
Following that, we draw on the discussion of commonsense knowledge in~\citet{liu2004conceptnet}\footnote{See Section 1.1 for more details.} and use it as a guideline for labeling categories beyond commonsense. 
We define the overall 45 categories beyond commonsense by asking three annotators to label 158 categories as commonsense or not, fulfilling a high agreement (Fleiss-$\kappa$ = 0.88). To ensure that the label sets under different noise sources do not overlap, we remove 9 categories also appearing in tail categories from the 45 categories and eventually receive 36 categories beyond commonsense. 
Lastly, the remaining 68 of the 158 classes are designated as the noise source SC. 



\begin{table}[]
	\small
	\centering
	\begin{tabular}{cccc}
		\toprule
		\multicolumn{2}{c}{\textbf{Operation}}                               & \textbf{Class} & \textbf{Size}      \\ \midrule
		\multicolumn{2}{c}{Crawling}                       & 3,334 & 1,407,306 \\ \midrule
		\multirow{5}{*}{\begin{tabular}[c]{@{}c@{}}Input \\ Filtering\end{tabular}} & Sample Length Filter & \multirow{4}{*}{3,334} & \multirow{4}{*}{777,135} \\ \cmidrule{2-2}
		& Format Normalization &                    &                    \\ \cmidrule{2-2}
		& Deduplication        &                    &                    \\ \cmidrule{2-2}
		& TF-IDF Filter        &                    &                    \\ \cmidrule{2-4} 
		& Sample Size Filter   & 736                  & 412,080                  \\ \midrule
		\multicolumn{2}{c}{Label Filtering}                   & 158   & 89,143     \\ \bottomrule
	\end{tabular}
	\caption{\label{filter_statistic}
		Statistics of reserved valid sample size and classes after different operations.
	}
\end{table}
\begin{figure}[t]
	\centering
	\includegraphics[width=0.95\columnwidth]{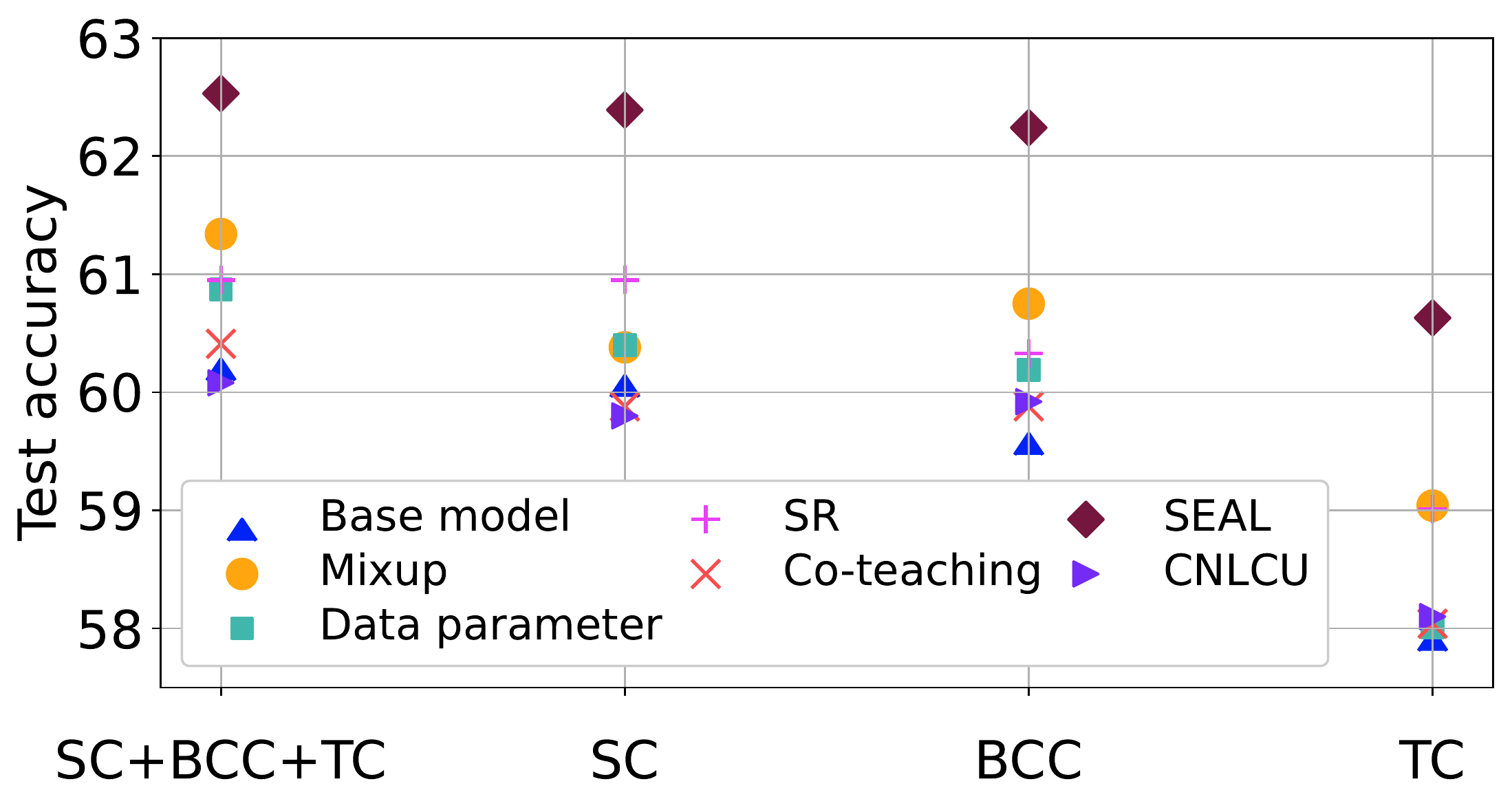}
	\caption{Test accuracy (\%) of typical LNL methods under distinct noise sources  with 10\% label noise.}
	\label{dif_noise_source_total}
\end{figure}
\begin{figure}[t]
	\centering
	\includegraphics[width=0.9\columnwidth]{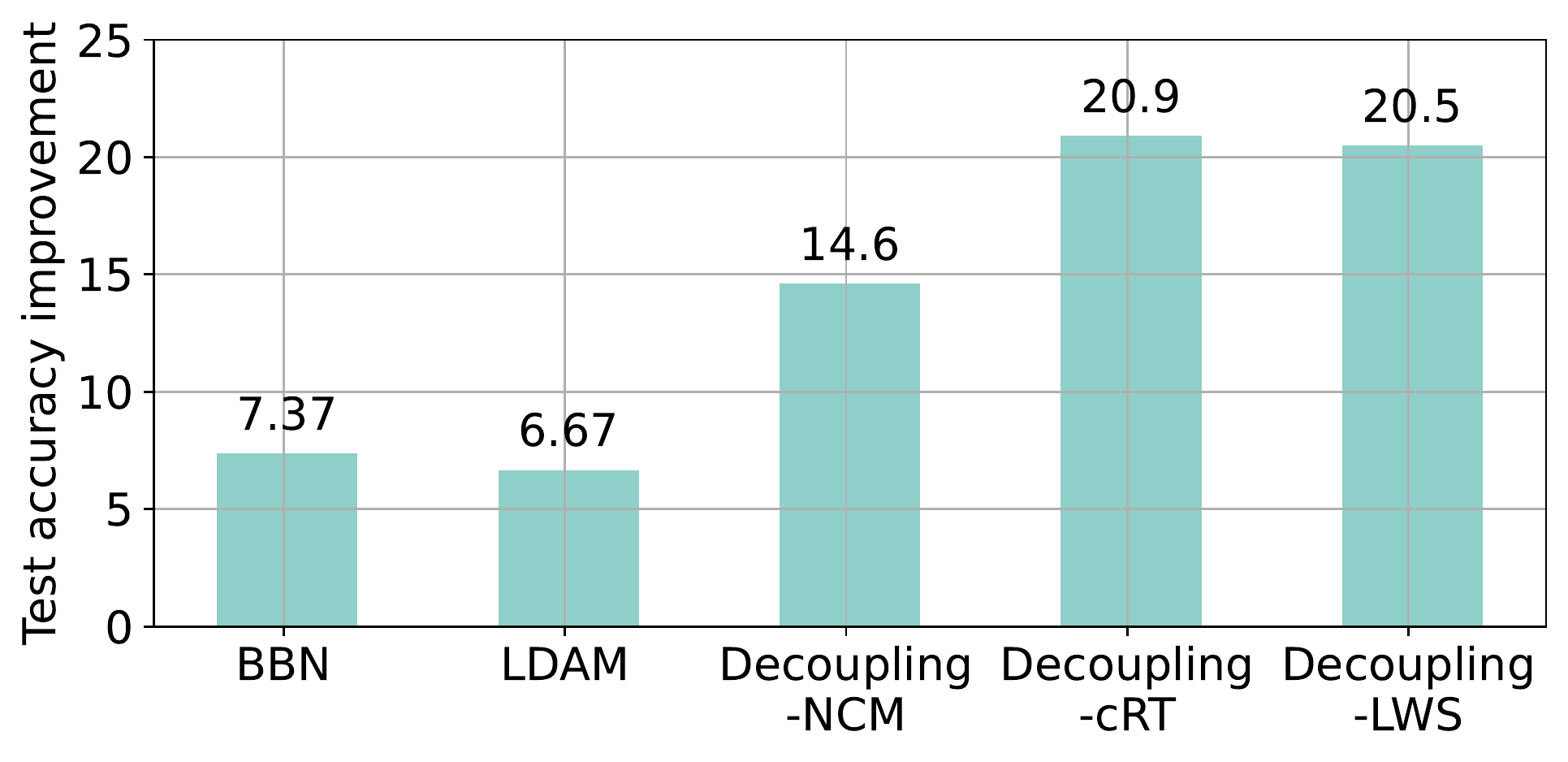}
	\caption{Performance improvements under different long-tailed learning methods in original papers.}
\label{ltl_ori}
\end{figure}

\section{Experiments Details}
\label{app:exp}
For each experimental dimension, we refine the hyperparameters for every baseline across different noise levels. Optimal hyperparameters are obtained by using a popular hyperparameter optimization tool \emph{Hyperopt}~\citep{bergstra2013making}.


\subsection{Effects of Distinct Noise Sources}
\label{app:exp_noise_source}
We examine the base model's performance under four different noise sources. In addition, Fig.~\ref{dif_noise_source_total} further compares the efficacy of typical LNL methods under various noise sources. We discover that regardless of the method employed, they are all less effective in reducing the effect of the noise source TC, further confirming our point of view.



\subsection{Long-tailed Distribution Properties}
\label{app:exp_lt}
\textbf{Baselines}:
Our training set follows a typical long-tailed class distribution akin to that in the real world. However, DNNs can be readily biased towards dominant classes with massive training data, triggering poor model performance on tail classes with limited data. This problem inspires large numbers of long-tailed learning studies.
To fully explore the characteristics of the NoisywikiHow dataset, we select five long-tailed learning methods in three classical categories as baselines:
(1) \emph{\textbf{BBN}}~\citep{zhou2020bbn}, which applies a resampling strategy to sample more tail-class samples for improving tail-class performance;
(2) \emph{\textbf{LDAM}}~\citep{cao2019learning}, which rebalances classes by designing an effective loss and training schedule;
(3) \emph{\textbf{Decoupling}}~\citep{kang2019decoupling}, which decouples the learning procedure (including three baselines: \emph{\textbf{Decoupling-NCM}}, \emph{\textbf{Decoupling-cRT}}, and \emph{\textbf{Decoupling-LWS}}) to understand how the long-tailed recognition ability is achieved. 
Complete experimental results of long-tailed learning methods are shown in Table~\ref{dim2}. We also demonstrate the settings of optimal hyperparameters in Table~\ref{hyper_LTL}.

\begin{figure*}[t]
\textsc{\centering
\includegraphics[width=2.0\columnwidth]{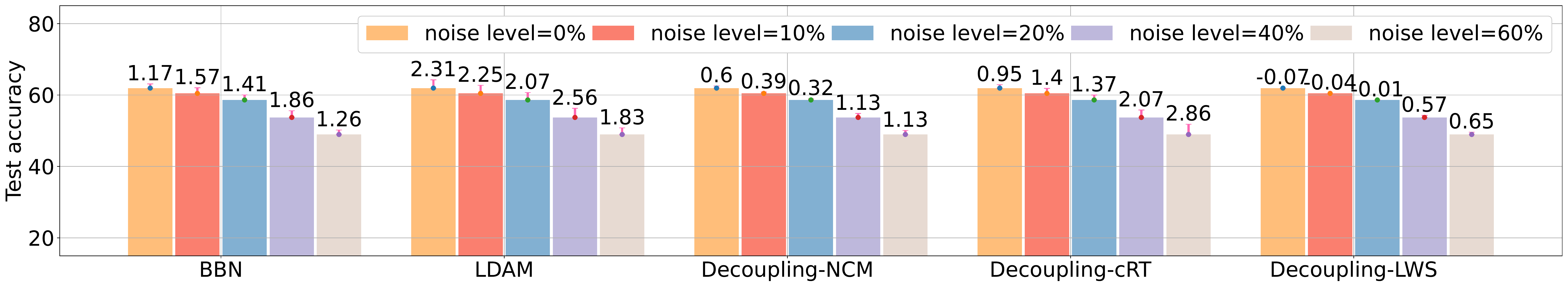}
\caption{Performance improvements over the base model under different long-tailed learning methods on NoisywikiHow. 
Given a method (e.g., BBN) and a noise level, a column height reflects performance when only using the base model. The length of the pink line on the column represents the performance boost from adopting the method.}
\label{ltl_nw}
}
\end{figure*}

\textbf{Results}:
We focus on the relative performance boost with various baselines in original papers and that on Noisywikihow. In Fig.~\ref{ltl_ori}, we find that all baselines evaluated on the CV datasets can address the long-tailed problem properly and achieve a significant test accuracy boost (7.37\%–20.9\%) in the original papers. 
However, as shown in Fig.~\ref{ltl_nw}, the performance improvements across varied noise levels on our NLP benchmark are limited, with some methods not exceeding the base model (-0.07\%–2.56\%). 



Experimental results indicate that the effectiveness of long-tailed learning methods needs to be examined on datasets with different modals. 
Moreover, although the base model obtains performance degradation with the increase in the noise level, the effectiveness of each long-tailed learning method is not significantly affected by the noise level variation. The main reason is that the test accuracy we report is the best peak accuracy, producing an effect similar to the early stop and thus preventing the model from overfitting label noise.

\begin{table*}[t]
\small
\centering
\begin{tabular}{cccccc}
\toprule
\multirow{3}{*}{\textbf{Method}} & \multicolumn{5}{c}{\textbf{Noise Level}}                                           \\ \cmidrule{2-6} 
& 0\%           & 10\%          & 20\%          & 40\%          & 60\%          \\ \cmidrule{2-6} 
& Top-1(Top-5)   & Top-1(Top-5)   & Top-1(Top-5)   & Top-1(Top-5)   & Top-1(Top-5)   \\ \midrule
Base model~\citep{lewis2020bart}&61.72(86.90)&60.28(85.92)&58.94(84.67)&54.57(82.38)&49.75(78.84) \\ 
Mixup~\citep{zhang2018mixup}&62.08(86.99)&61.34(86.61)&58.92(85.76)&55.91(83.49)&51.59(81.09) \\
Data Parameter~\citep{saxena2019data}&61.91(86.54)&60.87(86.24)&58.97(85.56)&54.70(82.05)&50.94(79.70) \\
SR~\citep{zhou2021learning}&62.32(87.35)&60.95(86.23)&58.78(86.09)&55.22(82.81)&50.14(79.70) \\
Co-teaching~\citep{BoHan2018CoteachingRT}&61.68(87.04)&60.41(86.11)&58.48(83.99)&54.57(81.58)&49.20(77.06) \\
CNLCU~\citep{xia2022sample}&61.25(86.67)&60.08(85.25)&58.33(83.52)&54.95(81.20)&50.91(78.08)\\
SEAL~\citep{chen2021beyond}&\textbf{63.29(87.65)}&\textbf{62.53(87.27)}&\textbf{61.49(86.57)}&\textbf{57.35(84.41)}&\textbf{52.73(81.56)} \\ \bottomrule
\end{tabular}
\caption{\label{dim3}
Top-1 (Top-5) classification accuracy (\%) of representative LNL methods on the test set of NoisywikiHow under different noise levels. 
Top-1 results are in bold.
}
\end{table*}

\begin{table*}[]
\small
\centering
\begin{tabular}{cc}
\toprule
\textbf{Method}         & \textbf{Optimal Hyperparameters Settings}              \\ \midrule
Mixup~\citep{zhang2018mixup}          & $\alpha=1$                                    \\
Data Parameter~\citep{saxena2019data} & lr\_inst\_param=0.2, wd\_inst\_param=0.0 \\
SR~\citep{zhou2021learning}             & $\tau=0.05, \lambda_0=0$, epochs=20           \\
Co-teaching~\citep{BoHan2018CoteachingRT}    & $T_k=8,\tau=\epsilon$ ($\epsilon$ is the noise level)        \\
CNLCU~\citep{xia2022sample} & $T_k=8,\tau_{min}=0.3$, fixed-length time intervals=5 \\
SEAL~\citep{chen2021beyond}          & Number of iterations=4                        \\ \bottomrule
\end{tabular}
\caption{\label{hyper_LNL1}
Optimal hyperparameter settings for different controlled real-world label noise on NoisywikiHow. 
}
\end{table*}


\begin{table*}[]
\small
\centering
\begin{tabular}{ccccc}
\toprule
\multirow{3}{*}{\textbf{Method}} & \multicolumn{4}{c}{\textbf{Noise Level}}                            \\ \cmidrule{2-5} 
& 10\%          & 20\%          & 40\%          & 60\%          \\ \cmidrule{2-5} 
& Top-1(Top-5)   & Top-1(Top-5)   & Top-1(Top-5)   & Top-1(Top-5)   \\ \midrule
Base model              & 59.83(85.66) & 58.89(84.97) & 55.66(82.03) & 51.49(78.29) \\
Mixup~\citep{zhang2018mixup}                  & 61.61(86.40) & 59.78(85.57) & 57.01(83.20) & 51.80(78.97) \\
Data Parameter~\citep{saxena2019data}          & 60.94(85.74) & 59.56(85.39) & 55.81(82.41) & 51.69(78.73) \\
SR~\citep{zhou2021learning}                      & 60.25(82.35) & 59.51(81.54) & 56.80(79.71) & 51.90(77.38) \\
Co-teaching~\citep{BoHan2018CoteachingRT}             & 60.86(86.06) & 59.97(85.16) & 56.92(82.99) & 52.95(79.85) \\
CNLCU~\citep{xia2022sample} & 60.46(85.84) & 59.49(84.92) & 57.16(83.05)  & 52.51(78.28)\\
SEAL~\citep{chen2021beyond}                    & \textbf{62.69(87.66)} & \textbf{61.41(86.99)} & \textbf{58.97(84.77)} & \textbf{54.66(80.92)} \\ \bottomrule
\end{tabular}
\caption{\label{dim5}
Top-1 (Top-5) test accuracy (\%) of representative LNL methods with controlled synthetic label noise.
}
\end{table*}

\begin{table*}[]
\small
\centering
\begin{tabular}{cc}
\toprule
\textbf{Method}         & \textbf{Optimal Hyperparameters Settings}              \\ \midrule
Mixup~\citep{zhang2018mixup}          & $\alpha=1$                                    \\
Data Parameter~\citep{saxena2019data} & lr\_inst\_param=0.2, wd\_inst\_param=0.0 \\
SR~\citep{zhou2021learning}             & $\tau=0.5, \lambda_0=0$, epochs=20           \\
Co-teaching~\citep{BoHan2018CoteachingRT}    & $T_k=3,\tau=\epsilon$  ($\epsilon$ is the noise level)     \\
CNLCU~\citep{xia2022sample} & $T_k=3,\tau_{min}=0.1$, fixed-length time intervals=5 \\
SEAL~\citep{chen2021beyond}          & Number of iterations=4                        \\ \bottomrule
\end{tabular}
\caption{\label{hyper_LNL2}
Optimal hyperparameter settings for different controlled synthetic label noise on NoisywikiHow.
}
\end{table*}

\begin{table*}[]
\small
\centering
\begin{tabular}{cccccc}
\toprule
\multirow{3}{*}{\textbf{Method}} & \multicolumn{5}{c}{\textbf{Noise Level}}                                           \\ \cmidrule{2-6} 
& 0\%           & 10\%          & 20\%          & 40\%          & 60\%          \\ \cmidrule{2-6} 
& Top-1(Top-5)   & Top-1(Top-5)   & Top-1(Top-5)   & Top-1(Top-5)   & Top-1(Top-5)   \\ \midrule
BBN~\citep{zhou2020bbn}                     & 63.11(\textbf{87.06}) & 62.03(\textbf{86.79}) & 60.03(\textbf{85.73}) & 55.59(\textbf{83.68}) & 50.22(80.47) \\
LDAM~\citep{cao2019learning}                & \textbf{64.25}(86.82) & \textbf{62.71}(86.19) & \textbf{60.69}(85.18) & \textbf{56.29}(82.53) & \textbf{50.79}(79.52) \\
Decoupling-NCM~\citep{kang2019decoupling}          & 62.54(85.59) & 60.85(85.61) & 58.94(84.71) & 54.86(82.58) & 50.09(79.76) \\
Decoupling-cRT~\citep{kang2019decoupling}          & 62.89(86.16) & 61.86(86.53) & 59.99(85.41) & 55.80(83.29) & 51.82(\textbf{81.40}) \\
Decoupling-LWS~\citep{kang2019decoupling}          & 61.87(85.75) & 60.42(85.96) & 58.61(84.63) & 54.30(82.20) & 49.61(79.50) \\ \bottomrule
\end{tabular}
\caption{\label{dim2}
Top-1 (Top-5) test accuracy (\%) of long-tailed learning methods on NoisywikiHow under different noise levels.
}
\end{table*}

\begin{table*}[]
\small
\centering
\begin{tabular}{cccccc}
\toprule
\multirow{2}{*}{\textbf{Method}} & \multicolumn{5}{c}{\textbf{Noise Level}}                               \\ \cmidrule{2-6} 
& 0\%         & 10\%        & 20\%       & 40\%       & 60\%        \\ \midrule
BBN~\citep{zhou2020bbn}                     & \multicolumn{5}{c}{-}                                        \\
LDAM~\citep{cao2019learning}                & C=0.2, s=10 & C=0.5, s=10 & C=0.7, s=7 & C=0.8, s=7 & C=0.8, s=10 \\
Decoupling-NCM~\citep{kang2019decoupling}          & \multicolumn{5}{c}{-}                                        \\
Decoupling-cRT~\citep{kang2019decoupling}          & \multicolumn{5}{c}{epoch'=5,   num\_samples\_cls=4}            \\
Decoupling-LWS~\citep{kang2019decoupling}          & \multicolumn{5}{c}{epoch'=5,   num\_samples\_cls=4}            \\ \bottomrule
\end{tabular}
\caption{\label{hyper_LTL}
Optimal hyperparameter settings for different controlled real-world label noise on NoisywikiHow.
}
\end{table*}

\end{document}